%% file: iclr2024_conference.tex
\documentclass{article} % For LaTeX2e
\usepackage{iclr2024_conference,times}

\input{math_commands.tex}

\usepackage{hyperref}
\usepackage{url}

\usepackage{threeparttable}
\usepackage{multirow}
\usepackage{graphicx}
\usepackage{makecell}
\usepackage{booktabs}
\usepackage{hyperref}
\usepackage{wrapfig}
\usepackage{changepage}
\usepackage{caption}
\usepackage{graphicx}
\usepackage{wrapfig}
\usepackage{color,soul}
\usepackage{tabularx}
\usepackage{enumitem}
\usepackage{tcolorbox}
\usepackage{listings}
\usepackage{tablefootnote}
\definecolor{green(pigment)}{rgb}{0.0, 0.65, 0.31}
\definecolor{green(munsell)}{rgb}{0.0, 0.66, 0.47}
\definecolor{green(ryb)}{rgb}{0.4, 0.69, 0.2}
\usepackage{fontawesome}
\newcommand \blfootnote[1]{
    \begingroup
        \renewcommand
        \thefootnote{}\footnote{#1}
        \addtocounter{footnote}{-1}
        \vspace{-1ex}
    \endgroup
}

\usepackage{xcolor,pifont}
\newcommand*\colourcheck[1]{%
  \expandafter\newcommand\csname #1check\endcsname{\textcolor{#1}{\ding{52}}}%
}
\colourcheck{blue}
\colourcheck{green}
\colourcheck{cyan}
\colourcheck{teal}
\colourcheck{red}
\newcommand{\xmark}{\textcolor{red}{\ding{55}}}%

\newcommand{\ferret}{Ferret }
\newcommand{\ferretns}{Ferret}

\title{Ferret: Refer and Ground Anything Anywhere at Any Granularity}

\author{$^\text{\faApple}$Haoxuan You\textsuperscript{1}\footnotemark[2]\>\,, Haotian Zhang\textsuperscript{2}\footnotemark[2]\>\,, Zhe Gan\textsuperscript{2}, Xianzhi Du\textsuperscript{2}, Bowen Zhang\textsuperscript{2}, Zirui Wang\textsuperscript{2}, \\
\textbf{Liangliang Cao\textsuperscript{2}, Shih-Fu Chang\textsuperscript{1}, Yinfei Yang\textsuperscript{2}} \\
\textsuperscript{1}Columbia University,  \textsuperscript{2}Apple AI/ML\\
\small{\texttt{haoxuan.you@cs.columbia.edu}, \texttt{\{haotian\_zhang2,zhe.gan,yinfeiy\}@apple.com}} \\
}

\iclrfinalcopy % Uncomment for camera-ready version, but NOT for submission.
\begin{document}

% \twocolumn[{%
% \renewcommand\twocolumn[1][]{#1}%
\maketitle
\begin{center}
    \vspace{-5mm}
    \centering
    \captionsetup{type=figure}
    \includegraphics[width=1\linewidth]{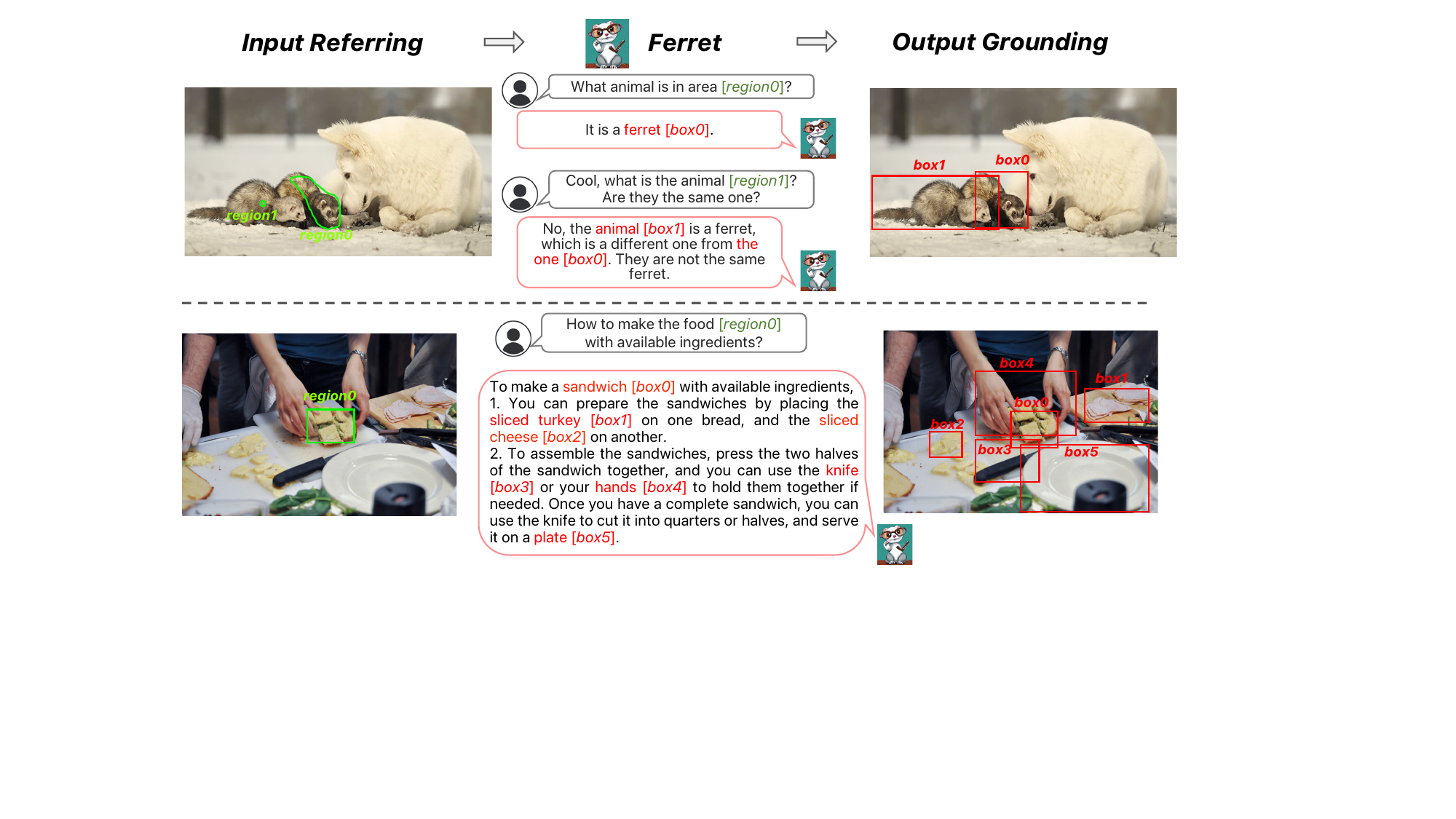}
    \vspace{-5mm}
    \captionof{figure}{\ferret enables \emph{referring} and \emph{grounding} capabilities for multimodal large language model (LLM). In terms of referring, a user can refer to a region or an object in point, box, or any free-form shape. The {\color{green(ryb)}\textit{regionN}} in the input will be replaced by the proposed hybrid representation before being fed into the LLM. In terms of grounding, \ferret is able to accurately ground any open-vocabulary descriptions. The {\color{red}\textit{boxN}} in the output denotes the predicted bounding box coordinates. 
    %\llcao{seems not good for Figure 1: (1) unclear why previous model cannot do? (2) too many details, the audience may not understand all of them}
    %\yyf{I actually like the examples here, but agreed with Liangliang it contains too many details. Maybe we only need the example 1 and 3, which are easier to understand. A proper figure caption is needed, too.}
    }
    \label{fig:intro}
\end{center}%
% }]

\maketitle

\begin{abstract}
\vspace{-2mm}
We introduce Ferret,
a new Multimodal Large Language Model~(MLLM) capable of understanding spatial referring of any shape or granularity within an image and accurately grounding open-vocabulary descriptions. 
To unify referring and grounding in the LLM paradigm, \ferret employs a novel and powerful hybrid region representation that integrates discrete coordinates and continuous features jointly to represent a region in the image. To extract the continuous features of versatile regions,  we propose a spatial-aware visual sampler, adept at handling varying sparsity across different shapes. Consequently, \ferret can accept diverse region inputs, such as points, bounding boxes, and free-form shapes. 
To bolster the desired capability of \ferretns, 
we curate GRIT, a comprehensive refer-and-ground instruction tuning dataset including 1.1M samples that contain rich hierarchical spatial knowledge, with 95K hard negative data to promote model robustness.
The resulting model not only achieves superior performance in classical referring and grounding tasks, but also greatly outperforms existing MLLMs in region-based and localization-demanded multimodal chatting. Our evaluations also reveal a significantly improved capability of describing image details and a remarkable alleviation in object hallucination. Code and data will be available at \textcolor{blue}{\href{https://github.com/apple/ml-ferret}{https://github.com/apple/ml-ferret}}.\blfootnote{$^\text{\faApple}$Work done during an internship at Apple. $^\dagger$Equal contribution.}
\end{abstract}

% \begin{figure}[t]
% \centering
% \includegraphics[width=1.05\linewidth]{figs/ferret_fig_spotlight.pdf}
% \caption{Overview of \ferret Model.}
% \label{fig:diagram}
% \end{figure}

\section{Introduction}
In vision-language learning, how to enable spatial understanding of models is a fundamental research problem.  Two desired capabilities stem from this problem: \emph{referring} and \emph{grounding}. Referring demands that the model can accurately comprehend the semantics of specific given regions~\citep{krahmer2012computational, kazemzadeh2014referitgame, mao2016generation, yu2016modeling, zellers2019recognition}, whereas grounding necessitates the model to localize the region in accordance with the given semantic description~\citep{luo2017comprehension, nagaraja2016modeling, yu2017joint, kamath2021mdetr}.

Essentially, \emph{referring} and \emph{grounding} demand the same type of knowledge: alignment of spatial information and semantics. Despite that, existing works mostly learn referring and grounding individually \citep{li2022grounded, wu2022grit, yu2017joint}. In comparison, humans can learn from one task and generalize the shared knowledge to the other task effortlessly, and are able to seamlessly integrate referring/grounding capabilities with daily dialogue and reasoning  \citep{zellers2019recognition}. 
% generalize the learned knowledge to both referring and grounding effortlessly.   
Inspired by the above gap, in this paper, we study three main questions: ($i$) How to unify referring and grounding in one framework, and will they benefit each other?  ($ii$) How to represent versatile types of regions that humans usually use for referring, such as point, box, scribble, and even free-form shapes? ($iii$) How to make referring and grounding open-vocabulary, instruction-following, and robust, which  are crucial for practical applications?

Targeting these three questions, we introduce \textbf{\ferretns}, a novel refer-and-ground Multimodal Large Language Model (MLLM). First of all, we choose MLLM as the bedrock of \ferret due to their powerful vision-language global understanding capability \citep{zhu2023minigpt, liu2023visual, li2023blip}. To unify referring and grounding, \ferret first represents the coordinates of regions in natural language numerical form,\footnote{Note that there is no additional vocabulary or position encoders introduced in \ferret model.} as illustrated in Figure~\ref{fig:diagram}. However, it is inefficient to use single point or box coordinates to represent versatile shapes of regions, such as strokes, scribbles, or complex polygons. These shapes are essential for more universal and precise human-model interaction. To solve this problem, we further propose a spatial-aware visual sampler to acquire the visual features for regions in any shape, taking care of the varying sparsity in those shapes. Then, the discrete coordinates and the continuous visual features are combined together to represent the visual regions in the input, composing a hybrid
region representation in \ferretns. Equipped with above methods, \ferret can deal with input that mixes referred regions with free-form text, and is able to ground the mentioned objects in its output by seamlessly generating the coordinates for each groundable object along with generating text.  
To our best knowledge, \ferret is the first work that is able to process free-formed region inputs in MLLMs.

In order to make the refer-and-ground capability in \ferret open-vocabulary, instruction-following, and
robust, we collect \textbf{GRIT}, a \textbf{G}round-and-\textbf{R}efer \textbf{I}nstruction-\textbf{T}uning dataset with 1.1M samples.  GRIT contains multiple levels of spatial knowledge, covering objects, relationships, region descriptions, and complex reasoning. It includes both text-in location-out (grounding) and location-in text-out (referring) data, as well as data that mixes location and text in both input and output.   The majority of the dataset is converted from existing vision(-language) tasks like object detection~\citep{krishna2017visual} and phrase grounding~\citep{yu2016modeling,plummer2015flickr30k} with carefully designed templates to make it instruction-following.
% To instantiate the versatile refer-and-ground capability, we further build a hierarchical and robust dataset, 
Additionally, 34K refer-and-ground instruction-tuning conversations are collected via the help of ChatGPT/GPT-4 \citep{openai2023gpt} to facilitate training an instruction-following and open-vocabulary refer-and-ground generalist. 
% \llcao{what are the advantages against previous datasets?} 
Moreover, we conduct spatial-aware negative data mining, which further promotes model robustness.
% \zhe{this paragraph is confusing, can we describe the datasets more clearly?}

\ferret subsumes strong open-vocabulary capabilities of spatial understanding and localization. When evaluated on conventional referring and grounding tasks, it achieves superior performance. More than that, we believe refer-and-ground capabilities should be integrated into daily conversations of humans, \textit{e.g.}, people refer to something they don't know and ask what it is used for (like Figure~\ref{fig:intro}). To evaluate this new capability, we introduce \textbf{\ferretns-Bench}, covering three new types of tasks: Referring Description, Referring Reasoning, and Grounding in Conversation. We benchmark existing MLLMs and observe that \ferret can outperform the best of them by 20.4\% on average. Moreover, \ferret demonstrates an intriguing property of alleviating object hallucinations.   

% It is evaluated not only on conventional referring and grounding datasets, but also multimodal dialogues that are based on regions and simultaneously require localization. Moreover, we found that \ferret demonstrates an intriguing property of alleviating object hallucinations.    

% Lastly, there are some concurrent works, Kosmos-2~\citep{peng2023kosmos} and Shikra~\citep{chen2023shikra}, studying enhancing the spatial understanding of MLLMs by also injecting point/box coordinates into the training data. \hl{Mark key differences.}
% that integrates discrete coordinates and continuous visual features jointly to represent a region,

In summary, our contributions are threefold. 
% (1). We enable fine-grained and open-vocabulary spatial understanding by referring and grounding in MLLM. 
($i$) We propose Ferret, that uses a hybrid region representation equipped with a novel spatial-aware visual sampler, to enable fine-grained and open-vocabulary referring and grounding in MLLM.
% allowing for reference to anything anywhere at any granularity as well as grounding in a .  \yyf{Are (1) and (2) basically the same thing ?}
($ii$) We construct GRIT, a large-scale ground-and-refer instruction tuning dataset, for model training. It also contains additional spatial negative samples to enhance model robustness.
%and a novel collection of refer-and-ground instruction tuning data. 
($iii$) We introduce \ferretns-Bench, to evaluate tasks jointly requiring referring/grounding, semantics, knowledge, and reasoning.  Our model exhibits superior performance in a wide range of tasks and reduces object hallucination. 

% \zhe{our eval set can also be a contribution.}
% \zhe{also, related work section is missing.}

% \yyf{How about add another contribution (4) here about extensive experiments and reducing hallucinations? }

% Novelty:

% 1. Enable fine-grained and open-vocabulary spatial understanding by referring and grounding in LLM.

% 2. Hybrid Referring Representation

% 3. Robust \& Hierarchical Data Collecting

% \section{Related Work}

\begin{table}[t]
\begin{center}
\begin{threeparttable}
\setlength{\tabcolsep}{3.5pt}
\fontsize{8.5}{9}\selectfont
\caption{Comparison of Ferret \textit{v.s.} recent MLLMs integrating spatial awareness. `Convention' refers to a comprehensive collection of publicly available data that has been transformed using templates, `GPT-Generate' signifies the generated refer/ground datasets employing GPT, and `Robustness' denotes datasets aimed at mitigating hallucination and improving robustness. Section~\ref{sec:grit} explains more details about each.}
\vspace{-2mm}
% \resizebox{0.95\textwidth}{!}{
\begin{tabular}{lcccccccc}
\toprule
% \multirow{2}{*}{Models}
\multirow{2}{*}{Model} & \multicolumn{3}{c}{Input Types} &  \multirow{2}{*}{\makecell[c]{Output \\ Grounding}} &  \multicolumn{3}{c}{Data Construction} &  \multirow{2}{*}{\makecell[c]{Quantatitive Eval.  \\ of Refer/Ground \\ w. Chat } } \\ [0.5ex] 
\cmidrule(r){2-4} \cmidrule(r){6-8}
% & Point & Box & Free-form  &  & Convention & GPT-Generate\tnote{\ding{169}}
% & Robustness &  \\
& Point & Box & Free-form  &  & Convention & GPT-Generate
& Robustness &  \\
% & Point & Box & Free-form  &  & C & G\tnote{\ding{169}}
% & R &  \\
% 768&209M&-    & Attn, FFN, LN1, LN2 & 31.85\\
\midrule
BuboGPT & \xmark & \xmark & \xmark & \cyancheck & \cyancheck & \xmark & \xmark & \xmark \\
Vision-LLM & \xmark & \xmark & \xmark & \cyancheck & \cyancheck & \xmark & \xmark & \xmark \\
% Kosmos-2 & \xmark & \cyancheck & \xmark & \cyancheck & \cyancheck & \xmark & \xmark & \xmark \\

Kosmos-2 & \xmark & \cyancheck & \xmark & \cyancheck &  \cyancheck & \xmark &  \xmark &  \xmark \\
Shikra & \cyancheck & \cyancheck & \xmark & \cyancheck & \cyancheck & \cyancheck & \xmark & \xmark\\
GPT4-ROI & \xmark & \cyancheck & \xmark & \xmark & \cyancheck & \xmark & \xmark & \xmark \\
PVIT & \xmark & \cyancheck & \xmark & \xmark & \cyancheck & \cyancheck & \xmark & \cyancheck \\
% \midrule
\textbf{\ferret} & \cyancheck & \cyancheck & \cyancheck & \cyancheck & \cyancheck & \cyancheck & \cyancheck & \cyancheck\\
\bottomrule
\end{tabular}
% \footnotetext{It means whether this work collects  new GPT-generated data. Directly using others' GPT-generated data doesn't count}
% }
% \vspace{-2mm}
\label{tab:related_work}
% \begin{tablenotes}
    % \item[\ding{169}] It means whether this work collects new GPT-generated \textbf{Refer-and-Ground}  data. 
% \end{tablenotes}
% \vspace{-4mm}
\end{threeparttable}
\end{center}
% \vspace{-2mm}
\end{table}

\section{Related Work}
\label{app:related_work}

\textbf{Multimodal large language models (MLLMs).}
Large Language Models (LLMs), including GPTs~\citep{brown2020language,gpt4}, PaLM~\citep{chowdhery2022palm}, BLOOM~\citep{scao2022bloom}, and LLaMA~\citep{touvron2023llama,touvron2023llama2}, have revolutionized research in NLP, spurring significant advances in multimodal language models as well. Early models primarily focused on large-scale image-text pre-training. Notable examples include SimVLM~\citep{wang2021simvlm}, GIT~\citep{wang2022git}, PaLI~\citep{chen2022pali}, PaLI-X~\citep{chen2023pali}, BLIP-2~\citep{li2023blip}, Flamingo~\citep{alayrac2022flamingo}, PaLM-E~\citep{driess2023palm}, CM3~\citep{aghajanyan2022cm3}, and CM3Leon~\citep{yu2023scaling}. Flamingo, in particular, pioneered the integration of a pre-trained CLIP image encoder with LLMs through gated cross-attention blocks, showcasing emergent multimodal in-context few-shot learning capabilities. Its open-sourced variants, such as OpenFlamingo~\citep{anas_awadalla_2023_7733589} and IDEFICS~\citep{laurenccon2023obelisc}, have garnered significant attention. Typically, these models undergo pre-training using millions or even billions of image-text pairs and interleaved image-text datasets~\citep{zhu2023multimodal}.

On the other hand, recent research has increasingly focused on using pre-trained LLMs for visual instruction tuning. Prominent examples include LLaVA~\citep{liu2023visual}, MiniGPT-4~\citep{zhu2023minigpt}, mPLUG-Owl~\citep{ye2023mplug}, Otter~\citep{li2023otter}, InstructBLIP~\citep{dai2023instructblip}, to name a few. In addition to text generation, recent models like FROMAGe~\citep{koh2023grounding}, GILL~\citep{koh2023generating}, Emu~\citep{sun2023generative}, have also enabled MLLMs for image retrieval and image generation. Please refer to Chapter 5 of \cite{li2023multimodal} for a detailed review.  

\textbf{MLLMs for referring and grounding.}
In the realm of existing literature, works such as Kosmos-2~\citep{peng2023kosmos} and Shikra~\citep{chen2023shikra},  closely resemble ours as they also enable MLLMs for fine-grained image comprehension and open-world referring and grounding. Additional works in this direction include  GPT4ROI~\citep{zhang2023gpt4roi}, PVIT~\citep{chen2023position}, BuboGPT~\citep{zhao2023bubogpt}, VisionLLM~\citep{wang2023visionllm}, and ContextDET~\citep{zang2023contextual}. Nevertheless, pivotal distinctions set our model apart.
First, prior endeavors supported only bounding boxes (and points in Shikra) as input.  Conversely, due to Ferret's innovative hybrid region representation, we accommodate a broader range of free-form shapes for referring, encompassing points, boxes, sketches, scribbles, polygons, and more.
Second, we meticulously curate an extensive refer-and-ground instruction tuning dataset.
Third, we introduce Ferret-Bench to facilitate forthcoming research and enhance evaluation benchmarks in this direction.
Lastly, our model exhibits superior performance compared to previous works, notably mitigating object hallucination to a significant extent. A more straightforward side-by-side comparison is shown in Tab. \ref{tab:related_work}.

\textbf{Unifying grounding and VL understanding.} Our work is also related to previous work that aims to unify text and bounding box output for vision-language (VL) models, such as UniTAB~\citep{yang2022unitab}, OFA~\citep{wang2022ofa}, and Unified-IO~\citep{lu2022unified}, which also represent bounding boxes using a set of additional discrete tokens as proposed in Pix2Seq~\citep{chen2021pix2seq,chen2022unified}. Ferret is unique in that ($i$) our model is built upon LLMs, marrying the power of LLMs and grounding, thus unlocking new capabilities such as grounded instruction tuning, and ($ii$) we handle bounding box coordinates as regular text tokens, avoiding the need for extra specialized tokens dedicated to representing boxes.

% \vspace{-1mm}
\section{Method}
% \vspace{-1mm}
We start with detailing the proposed hybrid region representation to depict regions of various shapes and formats. Then, we present the model architecture of \ferretns.

% \vspace{-1mm}
\subsection{Hybrid Region Representation}

\begin{wrapfigure}{r}{0.23\textwidth}
  \centering
  \captionsetup{font=footnotesize}
  \vspace{-5mm}
  \includegraphics[width=0.23\textwidth]{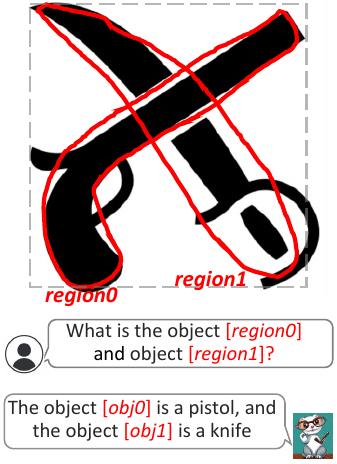}
  \vspace{-7mm}
  % \fontsize{4}{5}\selectfont
  \caption{Bounding box \textit{v.s.} Free-from Shape. These two objects have almost the same bounding box, causing ambiguity when relying on the box to refer to. Equipped with hybrid region representation, \ferret can separate them.}
  \label{fig:coordinate_drawback}
\vspace{-5mm}
\end{wrapfigure}

When referring to specific regions, three primary formats are generally used: point, box, and free-form shapes. While the point and box formats can be succinctly represented by coordinates (\emph{e.g.}, [$x, y$] for a point, and [$x_{\text{min}}, y_{\text{min}}, x_{\text{max}}, y_{\text{max}}$] for a box) as in \cite{peng2023kosmos, chen2023shikra}, the free-form shape is more versatile, encompassing a variety of region types such as scribbles, polygons, and masks. The advantage of free-form shape is straightforwardly illustrated in Figure~\ref{fig:coordinate_drawback}. Depicting free-form shapes through coordinates is computationally expensive and obscure, and its complexity hinders the model learning to establish a clear correlation between the provided coordinates and the corresponding regions.
% A straightforward example for illustration is shown in Fig. \ref{fig:coordinate_drawback}. 

% When users want to refer to some regions, there we define three main formats: point, box, and free-form shape. The free-form shape can cover multiple types of regions, such as scribble, polygons, and masks. Point and box can be easily represented by discrete coordinates \cite{kosmos2, shikra}, for example, use [$x, y$] to denote a point and [$x_{min}, y_{min}, x_{max}, y_{max}$] to denote a box.  However, expressing free-form shapes with coordinates is costly and obscure, which makes the model difficult to learn the correlation between given coordinates and corresponding regions, as shown in Fig. 1.  

\begin{figure}[t]
\centering
\makebox[\textwidth][c]{
\includegraphics[width=1.05\linewidth]{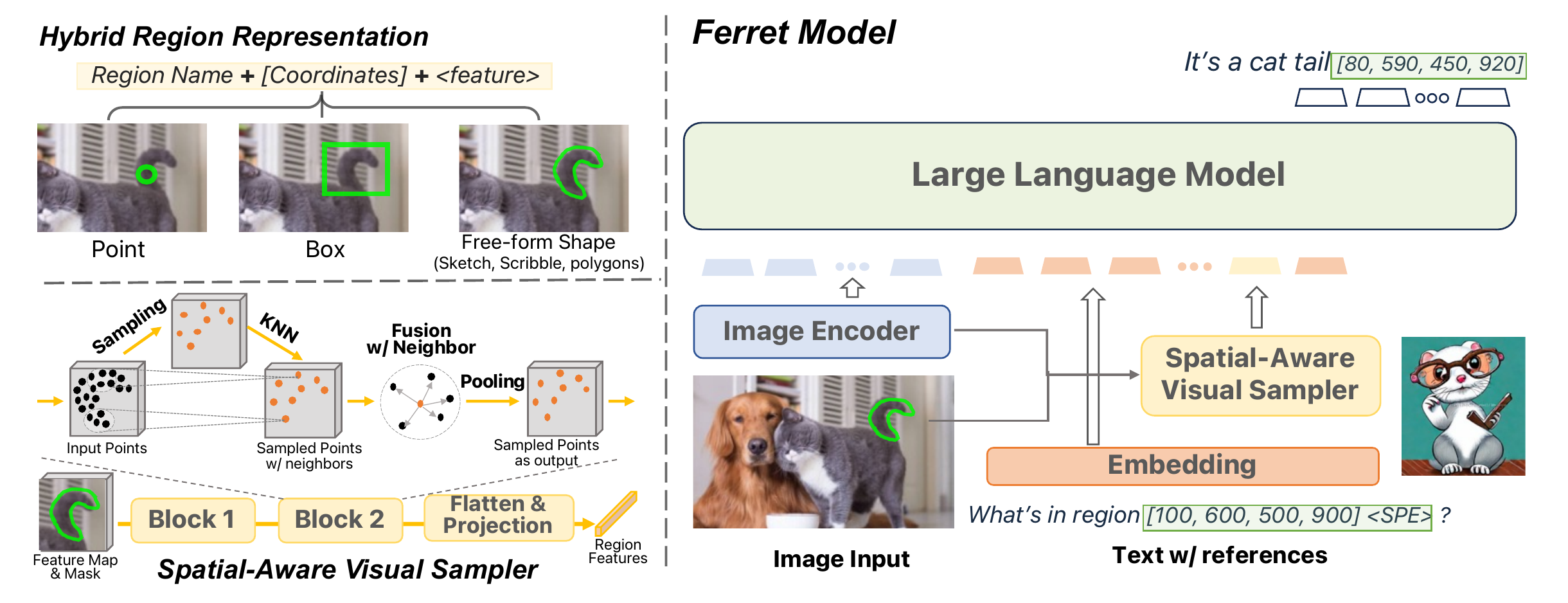}
}
\caption{Overview of the proposed \ferret model architecture. (Left) The proposed hybrid region representation and spatial-aware visual sampler. (Right) Overall model architecture. All parameters besides the image encoder are trainable.
}
% \vspace{-2mm}
\label{fig:diagram}
\end{figure}

To generalize across all three distinct formats, we propose a hybrid region representation that synergizes discrete coordinates with continuous visual features to refer to a particular region, which is shown in the top-left of Figure~\ref{fig:diagram}. For coordinates, following \cite{chen2021pix2seq, yang2022unitab}, we quantize each coordinate into one of the $n_{\text{bins}}$ discrete bins.\footnote{$n_{\text{bins}}=1000$ by default. The value is input invariant, which means for any input image size, the original coordinate will be mapped to the new coordinates. This makes the model robust to different input resolutions.} Regarding continuous visual features, for a given region $\mathbf{R}$, we first construct a 2D binary mask $\mathbf{M}$ of the same size as the image, marking a value of 1 inside the targeted region and 0 outside of the region. Then, the binary mask $\mathbf{M}$, jointly with the extracted image feature map $\mathbf{Z}$, is sent into our proposed spatial-aware visual sampler $s(\cdot)$, which will be detailed in Section~\ref{sec:architecture}, to extract the visual continuous feature $\mathbf{f}=s(\mathbf{M}, \mathbf{Z})$.

Finally, we represent a point with $\{x, y, \mathbf{f}_{R_p}\}$, where the region $R_p$ is a circle centered in $\{x, y\}$ with a fixed radius.\footnote{Radius is set to 5 by default.} A box or a free-form shape can both be represented by $\{x_{\text{min}}, y_{\text{min}}, x_{\text{max}}, y_{\text{max}}, \mathbf{f}_{R_{box}}\}$, where $x_{\text{min}}$/$x_{\text{max}}$ denotes the minimum/maximum $x$-axis coordinate of the region, and so forth for $y$-axis. $R_{box}$ denotes the input region.

% It's noted that in the model's output, we currently only predict the box coordinates instead of points or free-form shapes.

% \vspace{-2mm}
\subsection{Model Architecture}\label{sec:architecture}
% \vspace{-2mm}
As illustrated in Figure~\ref{fig:diagram}, \ferret is mainly composed of ($i$) an image encoder to extract image embeddings, ($ii$) the proposed spatial-aware visual sampler to extract regional continuous features, and ($iii$) an LLM to jointly model image, text, and region features.

\textbf{Input.} We feed the image into a pre-trained visual encoder, CLIP-ViT-L/14 ~\citep{radford2021learning}, to extract the image embeddings $\mathbf{Z}\in\mathbb{R}^{H\times W \times C}$. For text, we tokenize the text sequence using the pre-trained LLM's tokenizer and project them into text embeddings $\mathbf{T}\in\mathbb{R}^{L\times D}$. As for referred regions, we append the coordinates and a special token as a placeholder for continuous features after the name of the region: ``$\langle$region\_name$\rangle$ $\langle$coordinates$\rangle$ $\langle$SPE$\rangle$''. For example, ``a cat [100, 50, 200, 300]  $\langle$SPE$\rangle$''. If the name is unknown or hard to describe because multiple objects are included, we just use ``region'' or ``area'' as the ``$\langle$region\_name$\rangle$''. In this way, referred regions can be well mixed with ordinary texts to form complete sentences. 

% \textbf{Visual Sampler.}  Given extracted image feature map $\mathbf{Z}\in\mathbb{R}^{H\times W \times C}$ and the binary region mask $\mathbf{M}$, we introduced a visual sampler to extract the continuous visual feature of the region. We first project the extracted image feature $\mathbf{Z}$ to the same dimension of LLM by a linear layer, and then randomly sample $N$ positive points in $\mathbf{M}$. For each point, its feature is obtained by bilinear interpolation from the projected image feature map $\mathbf{Z'}$. And the $N$ point features are averagely pooled to obtain a single feature vector, which is used to replace the $\langle$SPE$\rangle$ token in the input.

\textbf{Spatial-aware visual sampler.}  
% Given extracted image feature map $\mathbf{Z}\in\mathbb{R}^{H\times W \times C}$ and the binary region mask $\mathbf{M}$, we introduced a spatial-aware visual sampler to extract the continuous visual feature of the region. 
The shape of the referred regions can be quite varied, not limited to just points or rectangle boxes. Grid-based processing like convolution or patch attention cannot handle irregular shapes. 
% \citet{zou2023segment}, they sample points inside the regions and average the point features, but it doesn't consider the shape information as well as the spatial correlation of points. 
Similar to our cases, 3D point clouds are also in irregular shape and show varied sparsity in the 3D space. Inspired by existing works in 3D point cloud learning \citep{qi2017pointnet, ma2022rethinking, wang2019dynamic}, we propose a spatial-aware visual sampler. 

Given extracted image feature map $\mathbf{Z}\in\mathbb{R}^{H\times W \times C}$ and the binary region mask $\mathbf{M}$, we first randomly sample $N$ positive points inside $\mathbf{M}$. For each point, its feature is obtained by bilinear interpolation. The $N$ points are fed into a cascade of blocks, where each of them includes three steps: sampling, gathering, pooling. (1) Sampling: $\frac{N}{r}$ points are sampled from $N$ points via farthest point sampling (FPS) algorithm \citep{qi2017pointnet++},\footnote{FPS starts from a random single point sampled from $N$ points. In each iteration, it samples one point from the rest points such that it is the farthest from the set of already sampled points. See detail in \cite{qi2017pointnet++}.} which can guarantee sufficient coverage. (2) Gathering: For each of the sampled points $x_i$, we search its $k$ nearest neighbors from the pool of previous $N$ points, and obtain a group of points $\{x_{i1}, x_{i2}, ..., x_{ik}\}$. Then, inspired by PointMLP \citep{ma2022rethinking}, for each group,  we fuse the features of sampled point $x_i$ and it neighbor points by:
\begin{equation}
\centering
\begin{aligned}
      h_{ik} &=  \mathbf{\sigma}([\mathbf{\theta}([\mathbf{Z}(x_{ik}) - \mathbf{Z}(x_{i}); C(x_{ik})-C(x_{i})]) ; \mathbf{Z}(x_{i}); C(x_{i})]) \,,
\end{aligned}
\end{equation}
where $x_{ik}$ is one of the neighbors of $x_i$, $\mathbf{Z}(x)$ denotes the point $x$'s feature (in the first block, it is interpolated from feature map $\mathbf{Z}$; in the succeeding blocks, it is the output feature from the previous block),  $C(x)$ denotes the 2D coordinates of point $x$, $[ ; ]$ means channel-wise concatenation of multiple vectors, $\mathbf{\theta}$ is implemented by a linear layer to adapt the relative local features, and $ \mathbf{\sigma}$ is also a linear layer to fuse each local feature from neighbors with sampled point feature. (3) Pooling: A max pooling is conducted to fuse $k$ neighbor features into one feature as the representation of the sampled point:
\begin{equation}
h_{i} = \max_{k: (x_{ik})\in \text{KNNs of } x_{i} } h_{ik}\,.
\end{equation}
After the three steps, we obtain fewer points but a more dense feature space since it incorporates the local neighbor features as well as their relative positions. In experiments, we set $N$=512, $r$=4 and $k$=24, and cascade two such blocks, which in the end outputs $32$ points with their features. Similar to ROIAlign \citep{he2017mask}, we flatten the point features into a single vector and project it to the dimension of LLM embeddings. The final feature is used to replace the $\langle$SPE$\rangle$ token in the input.

% We first project the extracted image feature $\mathbf{Z}$ to the same dimension of LLM by a linear layer, and then randomly sample $N$ positive points in $\mathbf{M}$. For each point, its feature is obtained by bilinear interpolation from the projected image feature map $\mathbf{Z'}$. And the $N$ point features are averagely pooled to obtain a single feature vector, which is used to replace the $\langle$SPE$\rangle$ token in the input.

\textbf{Output.} The above region denotations are used in Ferret input to refer to specific regions. In Ferret output, to achieve grounding, we generate the box coordinates right after the corresponding regions/nouns in the text response. For instance, ``There is a dog [100, 150, 300, 200] in the figure.'' With this data format, our model is expected to implicitly learn what is groundable in the current image and what their locations are. 

\textbf{LLM. } We consider Vicuna~\citep{chiang2023vicuna} as our language model, a decoder-only LLM~\citep{brown2020language} that is instruction-tuned on top of LLaMA~\citep{touvron2023llama}. Prior to being fed into the LLM, the image embeddings undergo transformation via an additional linear layer to match the embedding dimension of the text tokens.
% \vspace{-2mm}
\section{Grit: Ground-and-Refer Instruction-Tuning Dataset}
% \vspace{-2mm}
\label{sec:grit}
In this section, we present GRIT, a \textbf{G}round-and-\textbf{R}efer \textbf{I}nstruction-\textbf{T}uning dataset containing around 1.1M multimodal dialogues for model training. GRIT consists of three types of data: ($i$) public datasets that are converted into an instruction-following format (Section~\ref{sec:hierarchy}); ($ii$) instruction-tuning data generated via ChatGPT and GPT-4 (Section~\ref{sec:instruction_tuning}); and ($iii$) additional data from spatial negative mining for enhancing model robustness (Section~\ref{sec:negative_mining}). 

% \eat{
\begin{figure}[t]
\centering
\small
\includegraphics[width=1\textwidth]{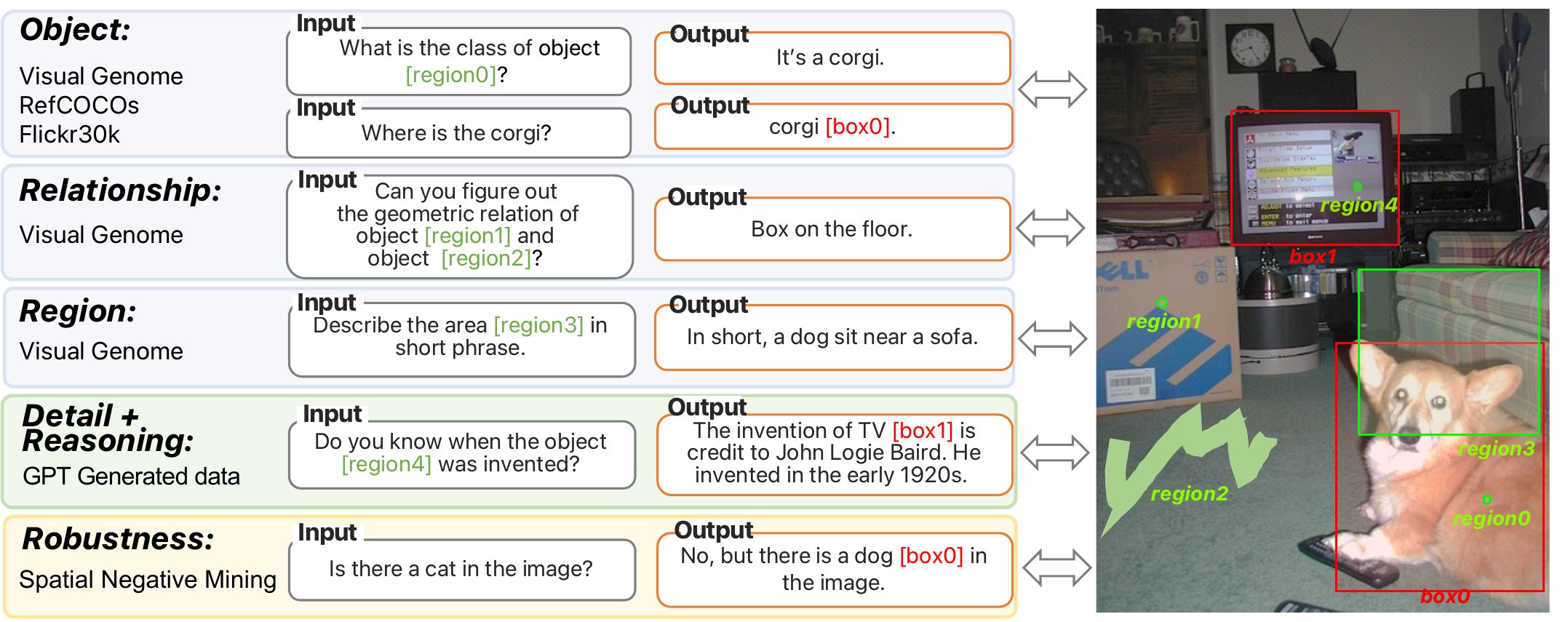}
\caption{\small Overview of the GRIT dataset for \ferret model training. It contains three types of data: ($i$) public datasets that are converted into an instruction-following format (the top-3 rows); ($ii$) data generated via prompting ChatGPT and GPT-4 (the 4th row); and ($iii$) negative data to enhance model robustness (the last row).
}
% \vspace{-4mm}
\label{fig:dataset}
\end{figure}
% } 

% \vspace{-2mm}
\subsection{Hierarchy} \label{sec:hierarchy}
% \vspace{-2mm}
Spatial understanding can be characterized by varying levels of granularity and task formats. During our dataset creation, we look into the following categories based on two dimensions:
\begin{itemize}[leftmargin=*]
% \vspace{-2mm}
    \item In terms of \textit{granularity}, we identify four main categories: 
    ($i$) individual objects, ($ii$) relationships among objects, ($iii$) descriptions of specific regions, and ($iv$) region-based complex reasoning.
    % \vspace{-1mm}
    \item In terms of \textit{task format}, we further divide the data into three distinct types: ($i$) Region-in Text-out data, ($ii$) Text-in Region-out data, and ($iii$) Text-Region combined data.\footnote{For Region-in Text-out data, the input highlights a specific region, prompting queries about it. For Text-in Region-out data, the input comprises textual descriptions, and the task is to pinpoint or ground the relevant region in its response. The combined Text-Region data integrates both text and region within a single sequence, which can be present in the input, output, or both.}
% \vspace{-1mm}
\end{itemize}

We compiled an extensive set of public data focusing on the aforementioned dimensions and converted them into an instruction-following format using carefully designed templates. A more in-depth view of these templates is available in Appendix~\ref{app:task_templates}.
% ~\footnote{\hl{Add detailed instruction template in appendix.}}.
% \haotian{shall we mention the quantity of each here?}

\textbf{Individual objects.} To achieve visual understanding at the object level, we select object detection datasets such as Visual Genome~\citep{krishna2017visual}, Object365~\citep{shao2019objects365}, and visual grounding datasets including RefCOCOs~\citep{yu2016modeling, lin2014microsoft, nagaraja2016modeling} and Flickr30k-Entities~\citep{plummer2015flickr30k}. The converted Visual Genome object data follow a \textit{Region-in Text-out} format. Additionally, to enable \ferret to understand free-form shapes, we apply SAM~\citep{kirillov2023segment} to Visual Genome object data to obtain a segmentation mask for each object, which is fed into the spatial-aware visual sampler to extract continuous region feature during training. The visual grounding datasets and Object365 data adhere to a \textit{Text-in Region-out} format. This section has in total 678k data.

\textbf{Relationships among objects \& descriptions of regions.} We selected data pertaining to object relationships and region captions from Visual Genome~\citep{krishna2017visual} to address these two facets, respectively. Both datasets employ a \textit{Region-in Text-out} format and 177k data are obtained. Similar to Visual Genome object data, we also extract segmentation masks of objects in Visual Genome relationship data via SAM.

\textbf{Region-based complex reasoning.} Regarding complex reasoning centered on specific regions, we constructed a novel dataset with the help of ChatGPT/GPT-4. It adopts a combined Text-Region format, and is detailed in the subsequent section. 

% \vspace{-2mm}
\begin{table*}[t]\centering
\caption{One example used in in-context learning to construct  GPT-Assisted Refer-and-Ground Instructon-Tuning. }
\begin{minipage}{0.99\columnwidth}\vspace{0mm}    \centering
\begin{tcolorbox} 
    \centering
     \hspace{-4mm}
      \scriptsize
    \begin{tabular}{p{0.99\columnwidth}}
    \VarSty{ {\bf Objects} } 
     \hspace{5.8cm} \multirow{5}{*}{ \includegraphics[height=5.2cm]{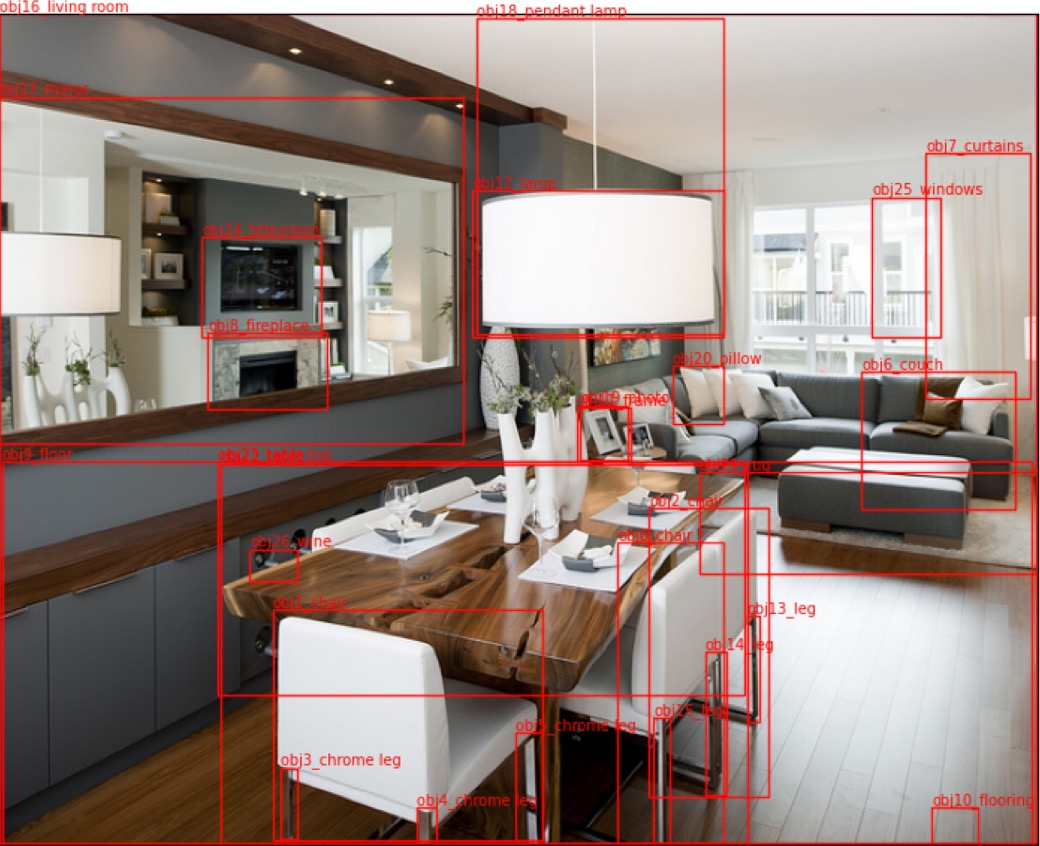} }
    Object 0: chair at [0.596, 0.637, 0.698, 0.997]. \\
    \hspace{0.5cm}... \\
     \VarSty{ {\bf Relationships} } \\
    Object 11 : frame $\rightarrow$ with $\rightarrow$ Object 9 : photo \\
    \hspace{0.5cm}... \\
    \VarSty{ {\bf Region Descriptions} } \\
    Region Description at [0.560, 0.466, 0.600, 0.529] : \\
    \hspace{1cm} a white picture frame with a black and white photo on it. \\
    \hspace{0.5cm}... \\
    \VarSty{ {\bf Global Caption} } \\
    White chairs sit around a polished wood dining table while \\
    \hspace{1cm} a sectional soft sits in the background. \\
    \hspace{0.5cm}... \\
\\
    % \midrule
   \VarSty{ {\bf Conversation} } \\
Question:  \\
What is the object [0.212, 0.539, 0.718, 0.820] made of? \\
=== \\
Answer: \\
The dining table over there is crafted from a tree trunk. \\
=== \\
Question: \\ 
Where would be a good place to have my dinner? Point out the locations in image.\\
=== \\
Answer: \\
I suggest having dinner at the table  [0.214, 0.541, 0.720, 0.997], possibly sitting on the white chair  [0.264, 0.717, 0.524, 0.995]. Also, enjoy the wine [0.242, 0.644, 0.288, 0.682]. \\
=== \\
Question: \\
In terms of comfort for resting, which object is better - the one [0.830, 0.431, 0.978, 0.596] or the one [0.264, 0.717, 0.524, 0.995]? \\
=== \\
Answer: \\
Among the two, the couch located [0.830, 0.431, 0.978, 0.596] would be more comfortable for resting. The object [0.264, 0.717, 0.524, 0.995] is a dining chair, which is typically less comfortable for extended relaxation. \\
.........
% === \\
% Question: \\
% Is the item located at [0.182, 0.238, 0.342, 0.461] real or is it a mirror reflection? \\
% === \\
% Answer: \\
% The television and the fireplace at [0.182, 0.238, 0.342, 0.461] are not real but are actually reflections in the mirror.  \\
% === \\
% Question: \\
% Can you describe the decorating style of the living room shown in this image? \\
% === \\
% Answer: \\
% The living room is decorated in a modern chic style. The image shows white chairs and lamps, a gray couch, and a wooden table and floor, all contributing to a modern aesthetic. The large mirror and built-in wine rack are also characteristic of modern design. Thus, the overall decor creates an atmosphere of elegance and sophistication, indicating a modern chic style.

    \end{tabular}
\end{tcolorbox}
\vspace{-2mm}
    \label{tab:example1_cap_conv_main}
\end{minipage}
\end{table*}

\subsection{GPT-assisted Visual Instruction Data Generation}
% \vspace{-2mm}
\label{sec:instruction_tuning}

% Inspired by the recent progress of LLMs \cite{gpt, chatGPT, LLAMA}, we collect the refer-and-ground instruction tuning data with the help of ChatGPT/GPT-4. 
Besides converting existing datasets by templates, dialogue instruction tuning data is proved to be critical for MLLM to understand human intention and generate fluent, natural, and long-form responses~\citep{liu2023visual, zhu2023minigpt, li2023m}. Few-shot prompting is widely used to obtain visual instruction tuning data, where textual scene descriptions of images and human-annotated dialogues are provided as few-shot demonstrations, and ChatGPT/GPT4 are prompted to generate new dialogue based on the new image's textual scene descriptions. 

However, previous instruction tuning data mainly focus on describing the entire image without explicitly specifying spatial-related information. To collect refer-and-ground instruction tuning data, we emphasize region-based spatial knowledge in the following three steps. ($i$) Besides objects and global captions usually used as before, our symbolic scene description additionally includes physical relationships between objects and region captions along with coordinates of them. ($ii$) In human-annotated dialogues, we add coordinates after the groundable regions or objects either in input or output or both, and the dialogues are typically focused on specific regions. It helps to implicitly prompt ChatGPT/GPT4 to follow similar patterns when generating new dialogues. A few-shot prompt used in our data is shown in Table~\ref{tab:example1_cap_conv_main}. ($iii$) The generated dialogues sometimes cannot follow the rules and patterns we wrote in system prompts and few-shot examples, which might be due to that the context of LLM input is too long to handle all the details. To alleviate it, we propose to use ChatGPT/GPT-4 again to refine the initially generated dialogues, whose context length is only 10\% of the data generated from the first round on average. 
% The prompts for refining are shown in xxx. 
To save cost, we use ChatGPT in the first round of generation and GPT-4 for refining.  34k dialogues in total are collected. 

Additionally, to exploit existing instruction-tuning data such as those in LLaVA~\citep{liu2023visual}, we apply an open-vocabulary object detector, GLIPv2 \citep{zhang2022glipv2}, on LLaVA-158k data to localize groundable nouns in the text. Then, we append the bounding boxes after the corresponding nouns, forming a pseudo-grounded LLaVA instruction data that are also used for training \ferretns. 
% \footnote{\hl{See Appendix for details.}}
% \vspace{-2mm}
\subsection{Spatial Negative Mining}\label{sec:negative_mining}
% \vspace{-2mm}

As highlighted in prior studies~\citep{li2023evaluating, liu2023aligning}, MLLM exhibits a propensity to hallucinate in response to yes/no questions. We observed a similar occurrence when inquiring about detailed regions. To address this, we also conduct negative sample mining by following two ways: ($i$) \textit{Image-conditioned Category Localization}, and ($ii$) \textit{Semantics-conditioned Category Localization}. They both ask the model to localize specific object categories, thereby enabling the model's ability to discern and potentially recognize the absence of certain objects.  They differ in how to select the negative category.  For ($i$), Object365 data are employed and we randomly select the object class from the vocabulary that is not shown in the given image.  For ($ii$), Flickr30k data are used and negative categories are sourced by utilizing ChatGPT/GPT4 to find entities that are most analogous to the original class, attribute, or quantity, \emph{e.g.}, `man' vs. `woman', `blue' vs. `yellow', `two' vs. `three'. 

We curate the data to maintain an equilibrium between positive and negative samples for each of the two types.\footnote{We observed that even though we don't collect other data specifically for training, \ferret demonstrates the capability to generalize robustness across diverse categories like relationships, events, \emph{etc}. We attribute this versatility to the potent compositional capabilities inherent to LLM.} 95k data are collected. A more comprehensive elaboration is provided in Appendix~\ref{app:negative_mining}. 

% for both referring and grounding, including ($i$) \textit{Region-centric Category Verification}, and ($ii$) \textit{Unconditional Category Localization}. For ($i$), this data type queries whether the object or region at a specific location pertains to a designated category. We choose categories of other objects within the same image as negative categories. For ($ii$), this type prompts the model to pinpoint specific categories, regardless of whether the queried category is actually present in the image, thereby testing the model's ability to discern and potentially recognize the absence of certain categories. 

% Negative categories are sourced by utilizing ChatGPT/GPT4 to pinpoint entities that are most analogous to the original class, attribute, or quantity, \emph{e.g.}, `man' vs. `woman', `blue' vs. `yellow', `two' vs. `three'. We curated a dataset that maintains an equilibrium between positive and negative samples for each of the two types.\footnote{We observed that even though we don't collect other data specifically for training, \ferret demonstrates the capability to generalize robustness across diverse categories like relationships, events, \emph{etc}. We attribute this versatility to the potent compositional capabilities inherent to LLM.} A more comprehensive elaboration is provided in Appendix~\ref{app:negative_mining}. 
% This resulted in the acquisition of 30k and 15k samples for each type, respectively

% \vspace{-2mm}
\section{Experiments}
% \vspace{-2mm}
First of all, we illustrate the training details of \ferretns.
Then in evaluation, we start with evaluating \ferret on conventional referring and grounding benchmarks (Sec. \ref{sec:refer} and \ref{sec:ground}). Then, we demonstrate the power of \ferret in more complex multimodal chatting with refer-and-ground capability in Sec. \ref{sec:gpt4}. 
For a detailed visualization of each, kindly check Appendix~\ref{app:viz}. We further ablate key components in \ferret (Sec. \ref{sec:ablation}), analyze the object hallucination of \ferret (Sec. \ref{sec:pope}) and discuss \ferret \textit{v.s.} GPT-4V (Sec. \ref{sec:analysis_gpt4v}).

\textbf{Training Details.} 
We initialize the image encoder with CLIP-ViT-L/14@336p, the LLM with Vicuna, and the projection layer with LLaVA's first-stage weights, leaving the visual sampler randomly initialized. After the initialization, \ferret is trained on the aforementioned GRIT data for three epochs, optimized by \citet{loshchilov2017decoupled} with a learning rate of $2e-5$ and a batch size of 128. The training takes $\sim$5/2.5 days on 8 A100 GPU for a \ferretns-13B/7B. During training, when input refers to regions, we randomly choose either the center points or the bounding boxes (or segmentation masks if available) to represent the regions. We perform de-duplication in training data to remove the samples that are in downstream evaluations. 

% \vspace{-2mm}
\subsection{Input Referring}
% \vspace{-2mm}
\label{sec:refer}

\begin{figure}[t]
    \centering
    % Left table
    \begin{minipage}{0.42\textwidth}
        \centering
        \setlength{\tabcolsep}{3pt}
        \fontsize{10}{9}\selectfont
         \captionof{table}{\small Results of referring object classification on three different referring types, including point, box, and free-form shape. `\ding{53}' means no such capability. }
        % \vspace{-2mm}
        \resizebox{1.0\textwidth}{!}{
        \begin{tabular}{lccc}
            \toprule
            % \multirow{2}{*}{Models}
            % & \multicolumn{3}{c}{Object Classification} \\
            % \midrule
            \multirow{2}{*}{Models} & \multicolumn{3}{c}{LVIS (\%)} \\[0.5ex]
            \cmidrule(r){2-4}
             &Point&Box & Free-form \\[0.5ex]
            % &&Description &Reasoning& Description &Reasoning&in Reasoning & All\\
            \midrule
            Random Guess & 50 & 50 & 50  \\[0.1ex]
            LLaVA & 50.1 & 50.3 & \ding{53} \\[0.1ex]
            \midrule
            % CLIP (ResNet50)* &87M& 31.3*\\
            Kosmos-2~\citep{peng2023kosmos}& \ding{53}& 60.25& \ding{53}\\[0.5ex]
            Shikra-7B~\citep{chen2023shikra}& 57.82& 67.71& \ding{53}\\[0.5ex]
            % 768&209M&-    & Attn, FFN, LN1, LN2 & 31.85\\
            GPT4-ROI~\citep{zhang2023gpt4roi}& \ding{53} & 61.76& \ding{53}\\[0.1ex]
            \midrule
            Ferret-7B& 67.94 &  79.42&  69.77 \\
            Ferret-13B&\textbf{68.35}& \textbf{80.46}  & \textbf{70.98}  \\
            % \hspace*{1em} w/ Early Specialization&129M& 35.18\\
            % \hspace*{1em} w/ Parallel Branch&129M& 34.18\\
            \bottomrule
        \end{tabular}
        }
        
        \label{tab:refer}
    \end{minipage}%
    \hfill
    % Right table
    \begin{minipage}{0.56\textwidth}
        \centering
        \setlength{\tabcolsep}{3pt}
        \fontsize{10}{9}\selectfont
        \captionof{table}{\small Results of grounded image captioning on the test set of Flickr30k Entities. BLEU@4, METEOR, CIDEr, and SPICE are used for the caption
evaluation. $F1_{all}$ and $F1_{loc}$ are used for grounding evaluation.  `--' means not reported. }
        % \vspace{-2mm}
        \resizebox{1.0\textwidth}{!}{
        \begin{tabular}{lcccc|cc}
            \toprule
            \multirow{2}{*}{Models} & \multicolumn{4}{c|}{Caption Eval.} & \multicolumn{2}{c}{Grounding Eval.} \\ [1ex]
            &B@4&M&C&S  & $F1_{all}$& $F1_{loc}$  \\[0.5ex]
            \midrule
            GVD~\citep{zhou2019grounded} & 27.3 & 22.5 & 62.3 & 16.5 & 7.55 & 22.2  \\ [0.5ex]
            Cyclical~\citep{ma2020learning} & 26.8 & 22.4 & 61.1 & 16.8 & 8.44 & 22.78  \\ [0.5ex]
            POS-SCAN~\citep{zhou2020more} & 30.1 & 22.6 & 69.3 & 16.8 & 7.17 & 17.49 \\ [0.5ex]
            UniTAB~\citep{yang2022unitab} & 30.1 & 23.7 & 69.7 & 17.4  & 12.95 & 34.79 \\ [0.5ex]
            Shikra-13B~\citep{chen2023shikra} & -- & -- & 73.9 & -- & -- & --  \\ [0.2ex]
            \midrule
            Ferret-7B & 35.1 & 24.6 & 74.8 & 18.0 & 15.02 & 37.62 \\ [0.5ex]
            Ferret-13B & \textbf{37.0} & \textbf{25.5} & \textbf{76.1} & \textbf{18.3} & \textbf{15.12} & \textbf{38.03} \\
            \bottomrule
        \end{tabular}
        }
    \label{tab:grounded_caption}
    \end{minipage}
    % \vspace{-2mm}
\end{figure}

% \begin{table}[t]
% \begin{wraptable}{r}{9cm} 
% \begin{center}
% % \setlength\extrarowheight{6pt}
% % \setlength{\tabcolsep}{3.5pt}
% \begin{tabular}{lccc}
% % \multicolumn{1}{c}{\bf PART}  &\multicolumn{1}{c}{\bf DESCRIPTION}
% \toprule
% % \multirow{2}{*}{Models}
% % & \multicolumn{3}{c}{Object Classification} \\
% % \midrule
% \multirow{2}{*}{Models} & \multicolumn{3}{c}{LVIS(\%)} \\[0.5ex]
% \cmidrule(r){2-4}
%  &Point&Box & Free-form  \\[0.5ex]
% % &&Description &Reasoning& Description &Reasoning&in Reasoning & All\\
% \midrule
% LLaVA & - & - & -  \\[0.1ex]
% \midrule
% % CLIP (ResNet50)* &87M& 31.3*\\
% Kosmos-2& -& 60.25& -\\[0.5ex]
% Shikra-7B& 57.93& 66.07& -\\[0.5ex]
% % 768&209M&-    & Attn, FFN, LN1, LN2 & 31.85\\
% GPT4-ROI& -& 70.21& -\\[0.1ex]
% \midrule
% Ours-7B& &  &   \\

% Ours-13B&\textbf{64.72}& \textbf{77.32}  & \textbf{72.53}  \\
% % \hspace*{1em} w/ Early Specialization&129M& 35.18\\
% % \hspace*{1em} w/ Parallel Branch&129M& 34.18\\
% \bottomrule
% \end{tabular}
% \caption{Result of Referring Object Classification.}
% \label{tab:refer}
% \end{center}
% \end{wraptable}
% \end{table}

The model's capability of understanding referring is reflected in that, given a referred region in the question, how accurately the model can understand the semantics of the referred region. To measure it, we start with the most basic semantics, \emph{object}, as it is fundamental and clear to define. To be more specific, the task we evaluate on is \textbf{\textit{Referring Object Classification}}: the question refers to a specific region in the image, and the model needs to classify the object in the region. Since \ferret and MLLMs usually generate free-form text responses, it is inaccurate to match the predicted class with the ground-truth class if directly asking the model to classify without constraints. Alternatively, we make it a binary-choice question in the format of ``Is the object $\langle$location$\rangle$ a $\langle$class\_A$\rangle$ or a $\langle$class\_B$\rangle$?''. We feed the binary-choice question and image into the MLLMs to obtain the response, and then detect if the response matches the ground-truth (GT) class by some rule.\footnote{Sometimes both GT class and negative class appear in the answer, \textit{e.g.}, ``The object is $\langle$class\_GT$\rangle$, not $\langle$class\_Neg$\rangle$''. Our rule removes the substring in-between ``not'' and comma/period, and then detects GT class.} 

To prepare the data, we used the validation split of LVIS dataset \citep{gupta2019lvis} covering over 1000
object categories, and sampled 2667 objects as the GT objects. Then, we randomly choose a different object category in the same image whose central point is close to the GT object as the negative object, and replace $\langle$class\_A$\rangle$ and $\langle$class\_B$\rangle$ with those two randomly to form 2667 questions. Additionally, to mimic the versatility of referring in human life, we replace the $\langle$location$\rangle$ with three different types: point, box, and free-form shape. For point, we randomly sample a point inside the GT object that is also near the GT object's boundary. For box, we use the GT bounding box provided by LVIS. For the free-form shape, we randomly generate some strokes inside the GT object to simulate that. Results on all three types of referring are summarized in Table~\ref{tab:refer}. \ferret can significantly outperform previous models \citep{peng2023kosmos, chen2023shikra} and handle all types of referring, a capability notably absent in previous works. We visualize some examples in Figure~\ref{fig:demo}. 

% \vspace{-2mm}
\begin{table}[t]
\begin{center}
\setlength{\tabcolsep}{3.5pt}
\caption{\small Performance comparison (Acc@0.5) on the referring expression comprehension (RefCOCO, RefCOCO+, RefCOCOg) and phrase grounding (Flickr30k Entities) tasks. $*$ indicates that the method is specifically fine-tuned in the second stage.}
% \vspace{-2mm}
\resizebox{0.95\textwidth}{!}{
\begin{tabular}{lccc|ccc|cc||cc}
\toprule
\multirow{2}{*}{Models} & \multicolumn{3}{c|}{RefCOCO} & \multicolumn{3}{c|}{RefCOCO+} & \multicolumn{2}{c||}{RefCOCOg} & \multicolumn{2}{c}{Flickr30k Entities}\\
&val&testA&testB  &val&testA&testB  &val&test & val&test \\
\midrule
MAttNet~\citep{yu2018mattnet} & 76.40 & 80.43 & 69.28 & 64.93 & 70.26 & 56.00 & 66.67 & 67.01 & -- & -- \\
OFA-L~\citep{wang2022ofa} & 79.96 & 83.67 & 76.39 & 68.29 & 76.00 & 61.75 & 67.57 & 67.58 & -- & -- \\
TransVG~\citep{deng2021transvg} & 81.02 & 82.72 & 78.35 & 64.82 & 70.70 & 56.94 & 68.67 & 67.73 & -- & 79.10 \\
%ViLBERT & -- & -- & -- & 72.34 & 78.53 & 62.61 & -- & -- & -- & -- \\
UNITER~\citep{chen2020uniter} & 81.41 & 87.04 & 74.17 & 75.90 & 81.45 & 66.70 & 74.02 & 68.67 & -- & -- \\
VILLA~\citep{gan2020large} & 82.39 & 87.48 & 74.84 & 76.17 & 81.54 & 66.84 & 76.18 & 76.71 & -- & -- \\
UniTAB~\citep{yang2022unitab} & 86.32 & 88.84 & 80.61 & 78.70 & 83.22 &  69.48 & 79.96 & 79.97 & 78.76 & 79.58\\
MDETR~\citep{kamath2021mdetr} & 86.75 & 89.58 & 81.41 & 79.52 & 84.09 & 70.62 & 81.64 & 80.89 & 82.3* & 83.8*\\
\midrule
Shikra-7B~\citep{chen2023shikra} & 87.01 & 90.61 & 80.24 & \textbf{81.60} & 87.36 & 72.12 & 82.27 & 82.19 & 75.84 & 76.54 \\
Ferret-7B & \textbf{87.49} & \textbf{91.35} & \textbf{82.45} & 80.78 & \textbf{87.38} & \textbf{73.14} & \textbf{83.93} & \textbf{84.76} & \textbf{80.39} & \textbf{82.21} \\
\midrule
Shikra-13B~\citep{chen2023shikra}& 87.83 & 91.11 & 81.81 & \textbf{82.89} & 87.79 & 74.41 & 82.64 & 83.16 & 77.41 & 78.44\\
Ferret-13B & \textbf{89.48} & \textbf{92.41} & \textbf{84.36} & 82.81 & \textbf{88.14} & \textbf{75.17} & \textbf{85.83} & \textbf{86.34} & \textbf{81.13} & \textbf{84.76} \\
\bottomrule
\end{tabular}
}
\label{tab:flickr_refcoco}
\end{center}
% \vspace{-5mm}
\end{table}

\subsection{Output Grounding}
% \vspace{-2mm}
\label{sec:ground}

\ferret performs well in referential dialogue, allowing for its integration into various VL tasks, notably those with grounding outputs. To rigorously assess the grounding capability, we first subject \ferret to benchmark visual grounding tasks in a generative paradigm. Then, to measure the alignments between words and regions, we further evaluate \ferret on grounded captioning task.

% By accessing \ferret on these tasks, our goal is to measure its proficiency in text-to-visual grounding, a cornerstone for VL models capable of integrating and interpreting diverse outputs with positions.

% The referring expression comprehension (REC) task asks the model to pinpoint the entity described by a textual referring expression in a given image. On the other hand, the phrase grounding task requires the model to determine a set of bounding boxes based on one or more related phrases tied to a distinct caption. By accessing \ferret on these tasks, our goal is to measure its proficiency in text-to-visual grounding, a cornerstone for VL models capable of integrating and interpreting diverse outputs with positions. For both tasks, we utilize uniform prompts, represented as ``\textit{What are the locations of $<$query$>$/$<$phrases$>$?}'', where \textit{$<$query$>$} denotes the textual referring expression, while \textit{$<$phrases$>$} stands for a ``comma-delimited'' aggregation of the given phrases. For text or phrases that incorporate a comma, we eliminate it during both training and inference to circumvent potential ambiguity. 

\begin{figure}[t]
\centering
\makebox[\textwidth][c]{
\includegraphics[width=1.05\linewidth]{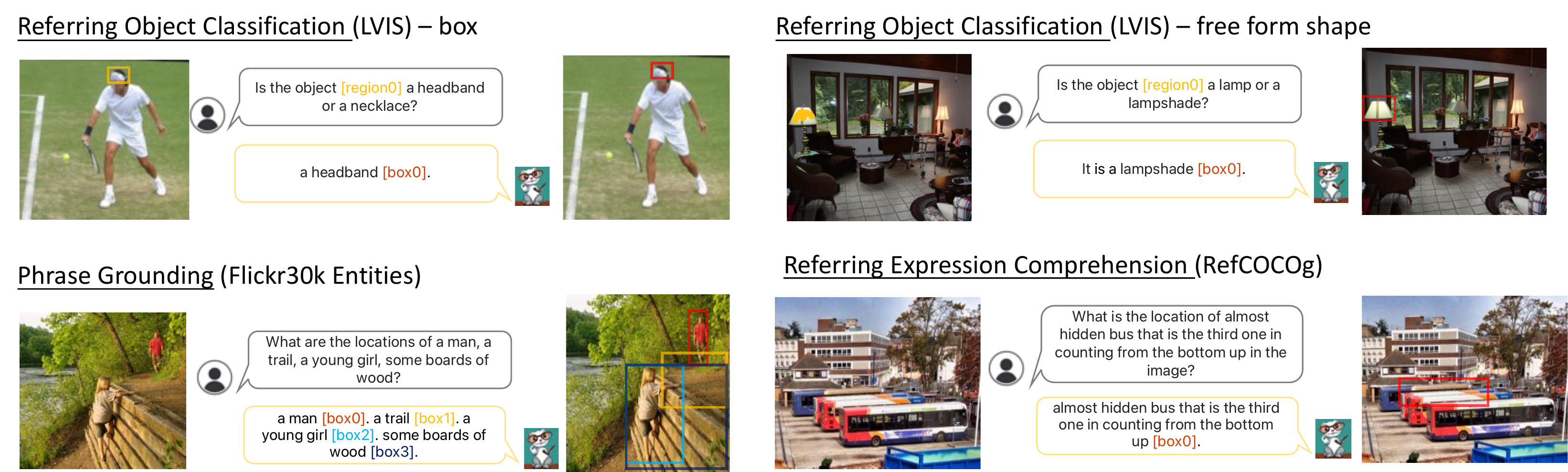}
}
\caption{Some examples demonstrating Ferret's referring and grounding capabilities. More visualizations are shown in Appendix~\ref{app:viz}.}
% \vspace{-2mm}
\label{fig:demo}
\end{figure}

\textbf{Visual grounding.} Visual grounding aims to ground language queries into
aligned image regions. We experiment on the sub-tasks of referring expression comprehension (REC) with three renowned benchmarks: RefCOCO~\citep{lin2014microsoft}, RefCOCO+~\citep{yu2016modeling}, and RefCOCOg ~\citep{mao2016generation}, and phrase grounding with Flickr30k Entities dataset~\citep{plummer2015flickr30k}. 
REC task involves a question or description about a specific area in an image, with the model expected to predict just one bounding box. Phrase grounding, conversely, seeks to associate all the noun phrases in the input sentence with corresponding boxes, requiring the model to predict these boxes and the word-box connections. For both tasks, we utilize uniform prompts, represented as ``\textit{What are the locations of $<$query$>$/$<$phrases$>$?}'', where \textit{$<$query$>$} denotes the textual referring expression, while \textit{$<$phrases$>$} stands for a ``comma-delimited'' aggregation of the given phrases. The model is trained to output in ``\textit{$<$query$>$ $[$box$]$.}'' format. The generated bounding box is considered correct if its intersection over union (IoU) with the GT box is greater than 0.5. As shown in Table~\ref{tab:flickr_refcoco}, \ferret achieves an outstanding performance on all metrics, and is comparable to specialized fine-tuning approaches~\citep{kamath2021mdetr}. 
% \hl{We have noticed a marginal decline in performance for 13B vs. 7B. We posit that certain datasets may exhibit noise due to their relatively small size.} 
Some results are visualized in Figure~\ref{fig:demo}.

\textbf{Grounded captioning.} The grounded captioning task requires the model to generate a caption and ground all generated noun phrases to image regions. 
The final predictions generally consist of three parts, \emph{i.e.}, the text caption, visual regions as boxes, and the grounding alignments between words and boxes. Following the established benchmarks on the Flickr30k Entities dataset, we evaluate captioning and grounding separately with the captioning metrics and grounding F1 scores, respectively. $F1_{all}$ evaluates grounding as a multi-label classification problem. We also report $F1_{loc}$ that only computes the grounding score on correctly predicted object words. 
Results are summarized in Table ~\ref{tab:grounded_caption}, and Ferret achieves state-of-the-art. 

% \vspace{-2mm}
\begin{table*}[t!]\centering
\caption{\small Visualization results of Referring Reasoning in Ferret-Bench to illustrate the difference between various models (LLaVA vs. Kosmos-2 vs. Shikra vs. Ferret (Ours)). For clarity, we have omitted the generated bounding box in the model's output. More visualizations can be found in Appendix~\ref{app:viz}}
\begin{minipage}{1.0\columnwidth}\vspace{0mm}    \centering
\begin{tcolorbox} 
    \centering
    %  \hspace{-10mm}
      \footnotesize
    \begin{tabular}{p{0.97\columnwidth} c}
   % \VarSty{ {\bf Input} } & \\
    \\
    \ArgSty{\bf Question:}  
& \hspace{-5.3cm} \multirow{4}{*}{ \includegraphics[height=3.8cm]{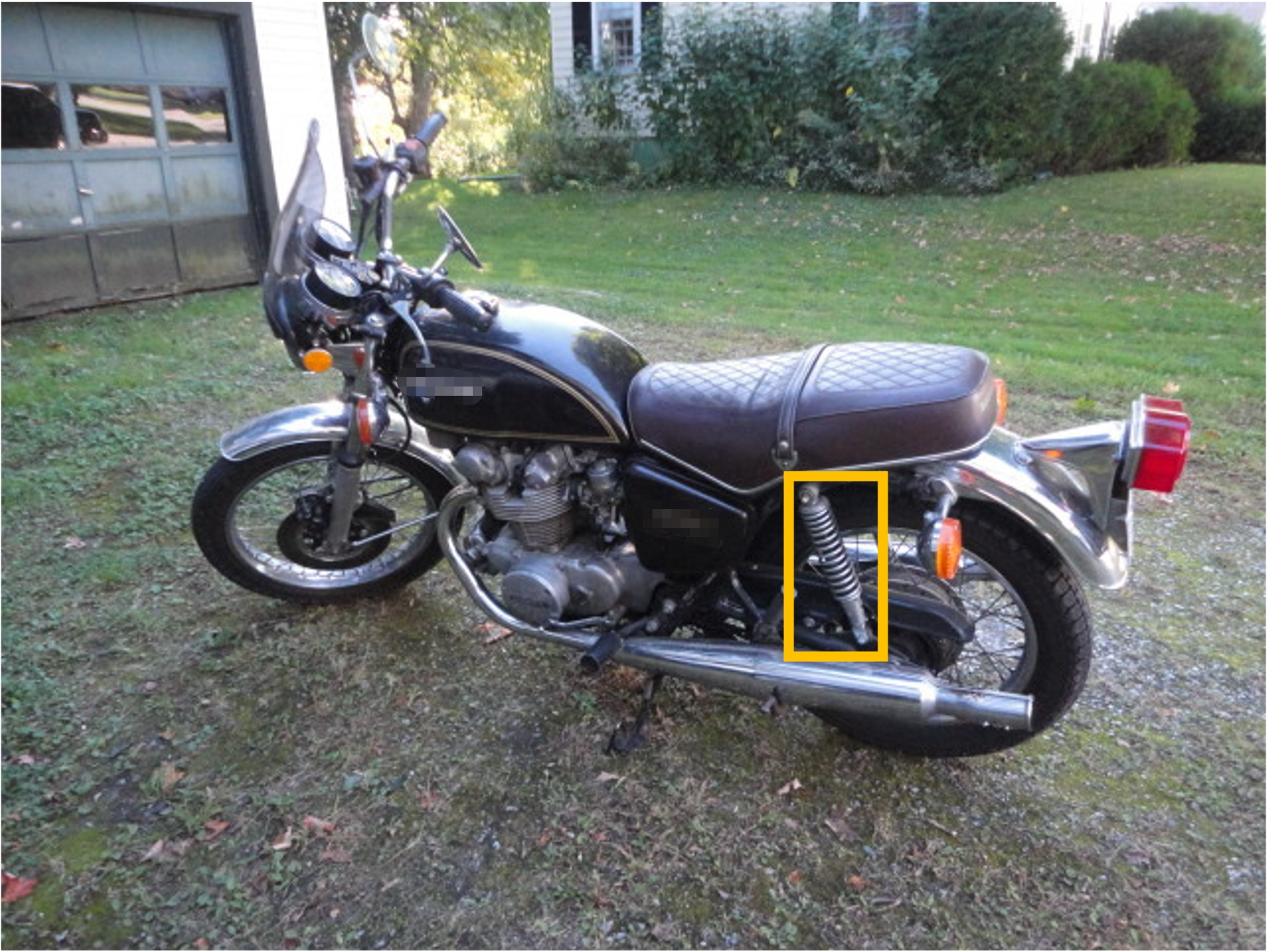} } \\
What is the purpose of the object \textcolor{orange}{[region0]} on the bike?  \\
\\
 % \\
\CommentSty{\bf Ground Truth:} \\ 
The object is a shock absorber on the bike. &\\ 
Its purpose is to absorb or dampen shock impulses.&\\  
It does this by converting the kinetic energy of &\\ 
the shock into another form of energy which is then &\\ 
dissipated. In the context of the motorcycle, &\\ 
it is particularly useful in providing comfort and &\\ 
stability for the rider, especially when travelling &\\ 
over uneven or rough terrains. \\
\\
\VarSty{ {\bf LLaVA:} } & \\
A black and silver Kawasaki motorcycle which is likely used for transportation, commuting, leisure rides, or long-distance travel. \\
   \VarSty{ {\bf Kosmos-2}: }& \\
The purpose of the object on the motorcycle is to store and protect the motorcycle's fuel tank. \\
   \VarSty{ {\bf Shikra}: }& \\
The answer is leg. \\
   \VarSty{ {\bf \ferret (Ours)}: }& \\
    The object is a shock absorber. This component is essential for the bike's stability, providing a smoother ride by soaking up the impact from bumps and uneven surfaces. It significantly improves the overall performance and comfort of the bike, enabling the rider to maintain control and stability even on rough terrains. \\
    \end{tabular}
\end{tcolorbox}
\vspace{-2mm}
    \label{tab:refer_reason_main}
\end{minipage}
\end{table*}

\subsection{\ferretns-Bench: Multimodal Chatting with Referring and Grounding}
% \vspace{-2mm}
\label{sec:gpt4}
Multimodal chatting has been an emergent ability of MLLMs. Previous benchmarks~\citep{liu2023visual} mainly evaluate conversation, detailed description, and complex reasoning via GPT-4 as a judge. Yet, a gap exists as no dataset currently evaluates multimodal chatting that necessitates referring or grounding actions, \textit{e.g.}, instances where individuals reference an unfamiliar object and inquire about its purpose. To benchmark this intriguing and practical capability, we introduce \ferretns-Bench that covers three kinds of region-based questions evaluating referring and grounding capability: ($i$) \textbf{Referring Description}: models are asked to describe a referred region based on \textit{its interaction with surrounding objects}. ($ii$) \textbf{Referring Reasoning}: models need to reason on top of one or more referred regions correctly. ($iii$) \textbf{Grounding in Conversation}: models are required to reason correctly and accurately ground/localize the objects/regions necessary for the reasoning. For the ease of benchmarking other methods, we represent the regions with boxes instead of points or free-form shapes. 

Specifically, we randomly sample 40 images from the COCO validation set for each type of question, and generate the questions and GPT-4's answers following the instruction generation pipeline in Sec. \ref{sec:instruction_tuning}. Following \citet{liu2023visual}, we feed the question and image into MLLMs to obtain the predicted answer, and then prompt GPT-4 to rate the predicted answer and pseudo answer from GPT-4 based on the ground-truth textual scene description (object, relationship, region caption, global caption). GPT-4 evaluates both the precision of referring understanding, object grounding, and correctness of semantics. The rating score ranges from 1 to 10, in which higher means better. We calculate the ratio of the predicted answer's score and the GPT-4 answer's score, which is then presented as a percentage to measure the performance of MLLMs. We also asked GPT-4 to give a comprehensive review for the rating and found that GPT-4 is good at measuring the degree of spatial precision, such as how much the predicted bounding box diverges from the GT box coordinate. We refer the readers to Appendix~\ref{app:ferretbench} for further elaboration. 

We use LLaVA-Bench~\citep{liu2023visual} and the proposed \ferretns-Bench to compare Ferret with previous models, including LLaVA~\citep{liu2023visual}, Shikra~\citep{chen2023shikra}, and Kosmos-2~\citep{peng2023kosmos}. Results are summarized in Table~\ref{tab:gpt4}. \ferret achieves superior performance in all types of tasks, boosting the score for the detailed description category from 68.3 to 80.9, and especially excels at the three new tasks demanding referring and grounding abilities. One visualization comparison is shown in Table~\ref{tab:refer_reason_main}, in which \ferret demonstrates strong spatial understanding and commonsense reasoning capability.  

\begin{table}[t]
\begin{center}
\setlength{\tabcolsep}{3.5pt}
\fontsize{9.5}{9}\selectfont
\caption{ Results on LLaVA-Bench and the proposed \ferretns-Bench via GPT4-as-a-Judge evaluation.}
% \vspace{-2mm}
\resizebox{0.92\textwidth}{!}{
\begin{tabular}{lcccc|cccc}
\toprule
% \multirow{2}{*}{Models}
&\multicolumn{4}{c|}{LLaVA-Bench} & \multicolumn{4}{c}{\ferretns-Bench}  \\
\cmidrule(r){2-9}
&\multirow{2}{*}{Conversation}&Detail & Complex &\multirow{2}{*}{Avg.}  & Referring & Referring  & Grounding  in &\multirow{2}{*}{Avg.} \\
&&Description &Reasoning& & Description &Reasoning& Conversation & \\
\midrule
% CLIP (ResNet50)* &87M& 31.3*\\
LLaVA\tablefootnote{The result on LLaVA-Bench is obtained by evaluating LLaVA released checkpoint. The slight discrepancy might be due to evolving GPT4 APIs. For Ferret-Bench, we employ the same conversation template as Ferret, providing LLaVA with a predefined input size, resizing all coordinates accordingly, and generating a response.}  &85.4 & 68.3 & 92.1 & 81.9 & 41.4 & 31.7 & 28.8 & 34.0 \\
% 768&209M&-    & Attn, FFN, LN1, LN2 & 31.85\\
\midrule
Kosmos-2 & 71.7 & 63.4 & 74.9 & 70.0 & 51.8 & 33.7 & 48.4 & 44.6\\
Shikra-7B & 80.6 & 70.7 & 88.1 & 79.9 & 46.0 & 41.6 & 50.1 & 45.9\\
% \midrule
Ferret-7B & 84.4 & 79.4  & 96.3 & 86.7 & 68.7 & 67.3 & 57.5 & 64.5 \\
Ferret-13B&\textbf{85.2}&  \textbf{80.9} & \textbf{96.4} & \textbf{87.5} & \textbf{70.6} & \textbf{68.7} & \textbf{59.7} & \textbf{66.3} \\
\bottomrule
\end{tabular}
}
% \vspace{-2mm}
\label{tab:gpt4}
\end{center}
% \vspace{-2mm}
\end{table}

%\section{Analysis \& Ablation}
%In this section, we ablate and analyze the behavior and design choice of the proposed \ferret model. 

% \textcolor{red}{\scriptsize(-2.6)}
\begin{figure}[t]
    \centering
    % Top-left table
    \begin{minipage}{0.46\textwidth}
        \centering
        \setlength{\tabcolsep}{3pt}
        \fontsize{9.5}{9}\selectfont
        \captionof{table}{Ablation study on the mutual benefit of grounding data and referring data.}
        % \vspace{-2mm}
        \resizebox{1.0\textwidth}{!}{
        \begin{tabular}{lcc|c}
            \midrule
            \multirow{2}{*}{Model} & \multicolumn{2}{c|}{Referring (LVIS)} & Grounding\\
            \cmidrule(r){2-4}
              & Point & Box &  Flickr30k \\
            \midrule
            \ferret &  \textbf{67.9} & \textbf{79.4} & \textbf{80.4} \\ [0.5ex]
            w/o Grounding data &  65.4  & 75.6 & \ding{53} \\ [0.5ex]
            w/o Referring data &  \ding{53} & \ding{53} & 79.8 \\
            % \midrule 
            \bottomrule
        \end{tabular}
        }
        \label{tab:ablate_mutual}
    \end{minipage}%
    \hfill
    % Top-right table
    \begin{minipage}{0.50\textwidth}
        \centering
        \setlength{\tabcolsep}{3pt}
        \fontsize{9.5}{9}\selectfont
        \captionof{table}{ Ablation study on the effectiveness of the proposed spatial-aware visual sampler. }
        % \vspace{-2mm}
        \resizebox{1.0\textwidth}{!}{
        \begin{tabular}{lccc}
            \midrule
            \multirow{2}{*}{Module} & \multicolumn{3}{c}{Referring (LVIS)} \\
            \cmidrule(r){2-4}
              & Point & Box &  Free-form  \\
            \midrule
            Spatial-aware Visual Sampler &  \textbf{67.9} & \textbf{79.4} & \textbf{69.8} \\ [0.5ex]
            Visual Sampler in SEEM &  67.1  & 77.2 & 68.9  \\ 
            % \midrule
            \bottomrule
        \end{tabular}
        }
        \label{tab:ablate_sampler}
    \end{minipage}
    
    \vspace{1pt} % Adjust the vertical spacing if necessary
    
    % % Bottom-left table
    % \begin{minipage}{0.46\textwidth}
    %     \centering
    %     \setlength{\tabcolsep}{3pt}
    %     \fontsize{9}{9}\selectfont
    %     \captionof{table}{\small Ablation study on the input image resolution of the CLIP image encoder.}
    %     \vspace{-2mm}
    %     \resizebox{1.0\textwidth}{!}{
    %     \begin{tabular}{lcc|c}
    %         \midrule
    %         \multirow{2}{*}{Model} & \multicolumn{2}{c|}{Referring (LVIS)} & Grounding\\
    %         \cmidrule(r){2-4}
    %           & Point & Box &  Flickr30k \\
    %         \midrule
    %         CLIP-ViT-L/336 &  \textbf{61.0} & \textbf{75.6} & \textbf{81.2} \\ [0.5ex]
    %         CLIP-ViT-L/224 &  59.4  & 73.6 & 79.5 \\
    %         % \midrule 
    %         \bottomrule
    %     \end{tabular}
    %     }
    %     \label{tab:ablate_res}
    % \end{minipage}%
    % \hfill
    % % Bottom-right table
    % \begin{minipage}{0.42\textwidth}
    %     \centering
    %     \setlength{\tabcolsep}{3pt}
    %     \fontsize{9}{9}\selectfont
    %     \vspace{-2mm}
    %     \captionof{table}{\small Ablation study on the model size of LLM being used.}
    %     \resizebox{1.0\textwidth}{!}{
    %     \begin{tabular}{lcc|c}
    %         \midrule
    %         \multirow{2}{*}{Model} & \multicolumn{2}{c|}{Referring (LVIS)} & Grounding\\
    %         \cmidrule(r){2-4}
    %           & Point & Box &  Flickr30k \\
    %         \midrule
    %         Vicuna-13B &  \textbf{61.0} & \textbf{75.6} & \textbf{81.2} \\ [0.5ex]
    %         Vicuna-7B &  60.1  & 74.2 & 80.3 \\
    %         % \midrule 
    %         \bottomrule
    %     \end{tabular}
    %     }
        % \label{tab:ablate_llm}
    % \end{minipage}
    % \vspace{-1mm}
\end{figure}

% \vspace{-2mm}
\subsection{Ablation}
\label{sec:ablation}
% \vspace{-2mm}
In the ablation studies below, in default, we ablate \ferretns-7B and mainly evaluate in referring object classification and grounding tasks on Flickr30k Entities validation set. 

\textbf{Mutual benefits of grounding and referring. } As shown in Table~\ref{tab:ablate_mutual}, grounding and referring, as two main capabilities emphasized in this paper, can actually benefit each other. Particularly, when adding grounding data into training, the referring performance gets improved, and vice versa. 
%The ablation result is shown in . 
 % Further, we observed that \ferret can alleviate object hallucination in Sec. \ref{sec:pope}. And, we ablate the main components of \ferret in Sec. \ref{sec:ablation}.

% \textbf{Hybrid region representation. }  Hybrid region representation enables joint versatile referring and grounding capabilities in a single model. We further study if the two cornerstones, coordinates and region features,  can be complementary in some common tasks compared with using only one of them. As shown in Tab. \ref{tab:ablate_hybrid}, the proposed hybrid representation can achieve better performance than solely using either one. 

% \textbf{Image resolution. }  The capability of referring and grounding has high demands to the strength of the visual backbone \citep{kamath2021mdetr, li2022grounded, shen2021much}. We further study if input resolution also matters.  We ablate CLIP-ViT-L/224  and CLIP-ViT-L/336 \citep{radford2021learning}, which have the same model structure and the number of parameters, but differ in the input image resolution. In Tab. \ref{tab:ablate_res}, we observed that higher resolution brings better performance, which might be attributed to more precise spatial understanding. 

\textbf{Spatial-aware Visual Sampler. } We ablate the effectiveness of the spatial-aware visual sampler by replacing it with the visual sampler in SEEM \citep{zou2023segment}, where they average the features of all the sampled points as the region feature. As we can see in Table~\ref{tab:ablate_sampler},  ours can outperform the previous visual sampler in all three referring tasks. 
% com  Hybrid region representation enables joint versatile referring and grounding capabilities in a single model. We further study if the two cornerstones, coordinates and region features,  can be complementary in some common tasks compared with using only one of them. As shown in Tab. \ref{tab:ablate_hybrid}, the proposed hybrid representation can achieve better performance than solely using either one.

\textbf{LLM model size. } We study how much LLM model size influences the performance of referring and grounding. As seen in Table~\ref{tab:refer}-\ref{tab:gpt4}, having a larger LM backbone can generally help.
% , \ref{tab:flickr_refcoco}, \ref{tab:grounded_caption}, 

% \vspace{-2mm}
\subsection{Object Hallucination}
% \vspace{-2mm}
\label{sec:pope}
Attribute to the incorporation of fine-grained spatial knowledge and negative mining, \ferret also exhibits strong power against the hallucination problem. We evaluate object hallucinations on the POPE benchmark \citep{li2023evaluating}. Results are summarized in Table~\ref{tab:pope}. \ferret has exhibited performance comparable to Shikra~\citep{chen2023shikra}, and far surpasses recent popular MLLMs.\footnote{Unlike other methods, \ferret refrains from relying on VQA. This decision stems from our observation that VQA answers tend to be concise, and this brevity can restrict the conversational capabilities of LLMs.} 

\begin{table}[t]
\begin{center}
\setlength{\tabcolsep}{3.5pt}
\caption{ Results on the object hallucination benchmark using the POPE evaluation pipeline~\citep{li2023evaluating}.}
% \vspace{-2mm}
\resizebox{0.92\textwidth}{!}{
\begin{tabular}{l|l|ccccccc}
% \multicolumn{1}{c}{\bf PART}  &\multicolumn{1}{c}{\bf DESCRIPTION}
\toprule
Datasets & Metrics & Ferret & Shikra & InstructBLIP & MiniGPT4 & LLaVA & MM-GPT & mPLUG-Owl  \\
\midrule
\multirow{5}{*}{Random}&Accuracy ($\uparrow$) & \textbf{90.24} & 86.90 & 88.57 & 79.67 & 50.37 & 50.10 & 53.97 \\
& Precision ($\uparrow$) & 97.72 & 94.40 & 84.09 & 78.24 & 50.19 & 50.05 & 52.07 \\
& Recall ($\uparrow$) & 83.00 & 79.26 &  95.13 & 82.20 & 99.13 & 100.00 & 99.60 \\
& F1 Score ($\uparrow$) & 89.76 & 86.19 & 89.27 & 80.17 & 66.64 & 66.71 & 68.39 \\
& Yes & 43.78 & 43.26 & 56.57 & 52.53 & 98.77 & 99.90 & 95.63 \\
\midrule
\multirow{5}{*}{Popular}&Accuracy ($\uparrow$) & \textbf{84.90} & 83.97 & 82.77 & 69.73 & 49.87 & 50.00 & 50.90 \\
& Precision ($\uparrow$) & 88.24 & 87.55 & 76.27 & 65.86 & 49.93 & 50.00 & 50.46 \\
& Recall ($\uparrow$) & 80.53 & 79.20 & 95.13 & 81.93 & 99.27 & 100.00 & 99.40 \\
& F1 Score ($\uparrow$) & 84.21 & 83.16 & 84.66 & 73.02 & 66.44 & 66.67 & 66.94 \\
& Yes & 45.63 & 45.23 & 62.37 & 62.20 & 99.40 & 100.00 & 98.57 \\
\midrule
\multirow{5}{*}{Adversarial}&Accuracy ($\uparrow$) & 82.36 & \textbf{83.10} & 72.10 & 65.17 & 49.70 & 50.00 & 50.67 \\
& Precision ($\uparrow$) & 83.60 & 85.60 & 65.13 & 61.19 & 49.85 & 50.00 & 50.34 \\
& Recall ($\uparrow$) & 80.53 & 79.60 & 95.13 & 82.93 & 99.07 & 100.00 & 99.33 \\
& F1 Score ($\uparrow$) & 82.00  & 82.49 & 77.32 & 70.42 & 66.32 & 66.67 & 66.82 \\
& Yes & 48.18 & 46.50 & 73.03 & 67.77 & 99.37 & 100.00 & 98.67\\
\bottomrule
\end{tabular}
}
\label{tab:pope}
\end{center}
% \vspace{-2mm}
\end{table}

% \subsection{Ablation}
% \label{sec:ablation}
% We ablate the following components of \ferret: hybrid representation, LLMs, and negative data mining. \haoxuan{TBD}

% \vspace{-2mm}
\begin{figure}[t]
\centering
\makebox[\textwidth][c]{
\includegraphics[width=1.1\linewidth]{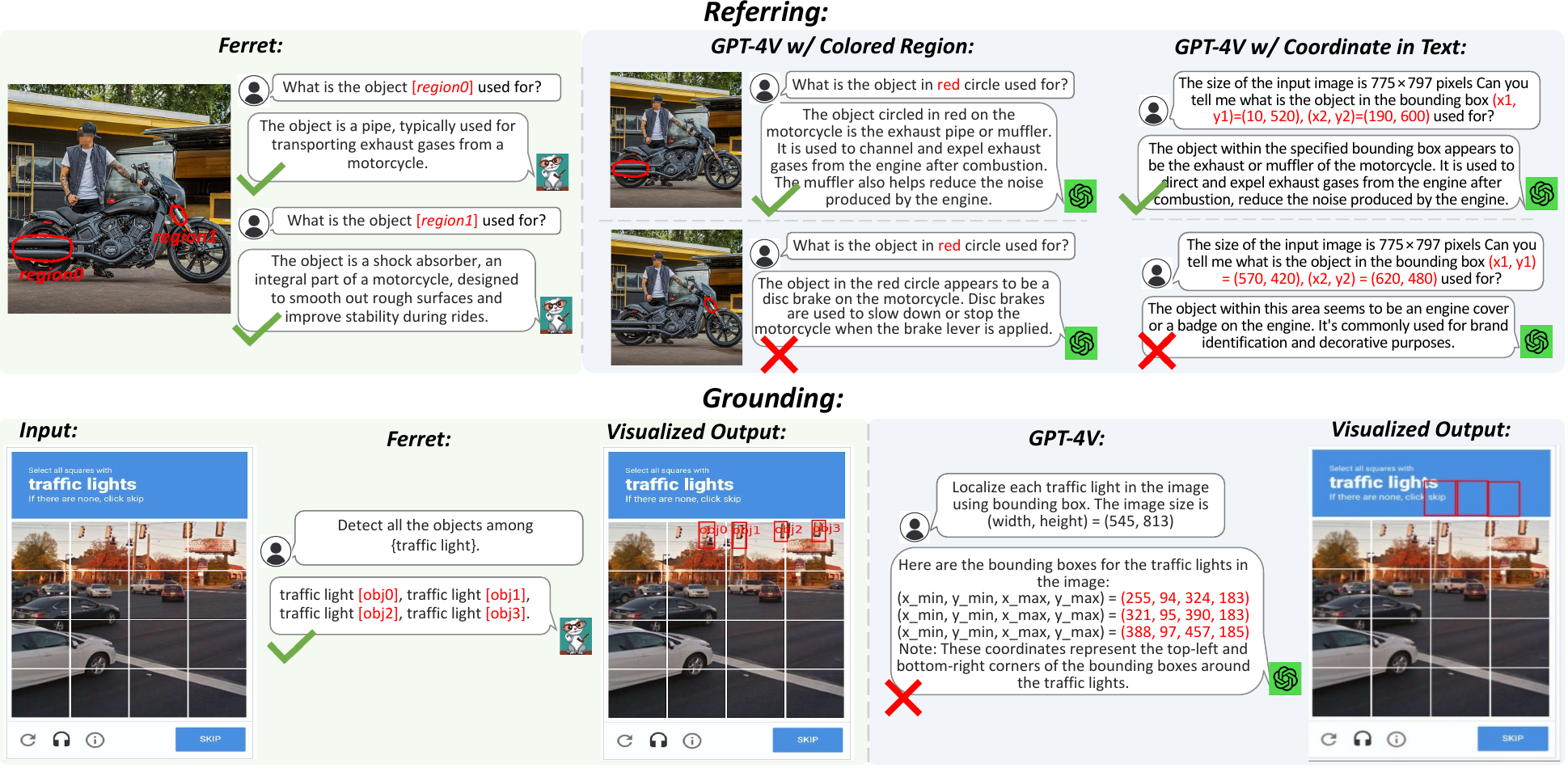}
}
\caption{\ferret \textit{v.s.} GPT-4V in Referring \& Grounding. In GPT-4V, we refer to regions by either marking them with red outlines in the image or adding their coordinates in the text. In terms of referring, we found that GPT-4V falls short in understanding relatively small regions. Similarly, for grounding, GPT-4V fails to localize relatively small objects in complex scenes. }
% \vspace{-2mm}
\label{fig:gpt4v}
\end{figure}

% \subsection{Discussion of \ferret \textit{v.s.} GPT-4V}
\subsection{\ferret \textit{v.s.} GPT-4V(ision): A Quick Glance at Referring \& Grounding}
\label{sec:analysis_gpt4v}
% \haotian{Topic: Ferret \& GPT4-V: A Deep Dive into Referring \& Grounding. }
Recently, GPT-4 released its multimodal version to the public, which is named GPT-4V. In a follow-up technical report~\citep{yang2023dawn}, GPT-4V's grounding ability is briefly touched. In this section, we use some examples to probe GPT-4V's referring and grounding capabilities, and compare with \ferretns.  For referring, GPT-4V is prompted with the following two ways: ($i$) referred regions are marked by red circle/outline in the image and the question asks about the region in red circle/outline. ($ii$) image is still but instead, we provide the image size and coordinates in question to refer to specific regions. As for grounding, we follow ~\citet{yang2023dawn}'s prompt, \textit{i.e.}, ``Localize $\langle$class$\rangle$ in the image using bounding boxes. The image size is (width, height)''. 

As we observed, GPT-4V is able to understand the referring to a certain extent via either colored region in the image or coordinates in text. However, compared with \ferretns, GPT-4V falls short in precise understanding when referring to small regions, \textit{e.g.}, the `shock absorber` in the motorcycle (see the upper example in Figure~\ref{fig:gpt4v}). On the other hand, GPT-4V is more knowledgable in commonsense, \textit{e.g.}, it can further highlight that the exhaust pipe can reduce the noise, a nuance potentially attributable to GPT-4's enhanced linguistic capabilities. In regard to grounding, we tested GPT-4V with CAPTCHAs, a task which is also mentioned in ~\citet{yang2023dawn}. In the traffic light example, Ferret excels at accurately identifying most traffic lights even in cluttered scenes, as demonstrated in the bottom example of Figure~\ref{fig:gpt4v}. 
% In a traffic light example, GPT-4V struggles to accurately pinpoint the traffic lights with a cluttered scene while \ferret can adeptly identify most of them (see the bottom example in Figure~\ref{fig:gpt4v}). 

That being said, in the realm of general question answering, GPT-4V is notably impressive. It skillfully manages not only initial questions but also follow-up inquiries linked to specific regions, delivering in-depth responses. Nevertheless, Ferret shines especially when precise bounding boxes for grounding are needed, and catering to those applications that require pinpoint accuracy in smaller regions. This is precisely where Ferret steps in to fill the gap.

% That being said, in the realm of general question answering, GPT-4V is notably impressive. It not only addresses questions but also handles follow-up queries related to a referring region, providing comprehensive answers. \hl{However, the model struggled referring to relatively small regions and was unable to produce accurate bounding boxes for grounding, indicating its current unsuitability for such applications. This is precisely where Ferret steps in to fill the gap.} 

\section{Conclusion}
% \vspace{-2mm}
We present Ferret, a new multimodal large language model adept at referring and grounding. Ferret can refer image regions in any free-form shape, and automatically establish grounding for text deemed groundable by the model. We have curated the GRIT dataset for model training, and the Ferret-Bench dataset for evaluation. Ferret, like most MLLMs, may produce harmful and counterfactual responses. For future work, inspired by LISA~\citep{lai2023lisa}, we plan to enhance Ferret to be able to output segmentation masks in addition to bounding boxes.

\bibliography{iclr2024_conference}
\bibliographystyle{iclr2024_conference}

\newpage
\appendix
\input{appendix}

\end{document}

%% file: math_commands.tex
\usepackage{amsmath,amsfonts,bm}

\usepackage{tcolorbox}

\usepackage{xcolor,amsmath}
\usepackage[linesnumbered,ruled,vlined]{algorithm2e}
\DontPrintSemicolon

\usepackage{xcolor}
\definecolor{mygreen}{HTML}{3cb44b}
\usepackage{varwidth}
\renewcommand{\ArgSty}[1]{\textnormal{\ttfamily #1}\unskip}
\SetKwComment{Comment}{\color{green!50!black}\# }{}
\renewcommand{\CommentSty}[1]{\textnormal{\ttfamily\color{green!50!black}#1}\unskip}

\newcommand{\var}{\texttt}

\SetKwProg{Function}{def}{:}{}

\SetKwProg{For}{for}{:}{}
\SetKwProg{If}{if}{:}{}
\newcommand{\VarSty}[1]{\textnormal{\ttfamily\color{blue!90!black}#1}\unskip}

\def\eqref#1{equation~\ref{#1}}
\def\1{\bm{1}}

\DeclareMathAlphabet{\mathsfit}{\encodingdefault}{\sfdefault}{m}{sl}
\SetMathAlphabet{\mathsfit}{bold}{\encodingdefault}{\sfdefault}{bx}{n}

%% file: appendix.tex
\section{Details of Dataset}
\label{app:dataset}

\subsection{Task Templates for Public Datasets} \label{app:task_templates}
In Section~\ref{sec:hierarchy}, we mentioned using carefully designed task templates to convert public datasets such as Visual Genome into instruction-following format. The task templates we used are provided in Table~\ref{tab:task_temp}. For simplicity, we only list three examples for each task. 

\begin{table*}[htbp]
\centering
\caption{Examples of task templates Ferret used to transfer different public data types into the instruction-following format.}
\resizebox{1.0\textwidth}{!}{%
\begin{tabular}{l|l}
\toprule
\textbf{Task} & 
\textbf{Three randomly chosen examples from many.}  \\
\cmidrule(lr){1-1}\cmidrule(lr){2-2}
\multirow{3}{*}{Referring-Object}
& What is the class of the object $<$location$>$ within the image?  \\
& Classify object $<$location$>$ in the image.    \\
& Identify the object $<$location$>$ in the image.  \\
\cmidrule(lr){1-1}\cmidrule(lr){2-2}
\multirow{3}{*}{Referring-Relation}
& What does $<$object1$>$  $<$location1$>$ do to $<$object2$>$  $<$location2$>$ of the image?  \\
& What is the physical relation between $<$object1$>$  $<$location1$>$ and $<$object2$>$  $<$location2$>$?    \\
& Can you figure out the geometric relation of the $<$object1$>$  $<$location1$>$  and $<$object2$>$  $<$location2$>$?  \\
\cmidrule(lr){1-1}\cmidrule(lr){2-2}
\multirow{3}{*}{Referring-Region}
& Describe the region $<$location$>$ in a short phrase.  \\
& What is in the region $<$location$>$? Describe in a phrase.    \\
& Capture in a phrase: what's near region $<$location$>$ in the picture?  \\
\cmidrule(lr){1-1}\cmidrule(lr){2-2}
\multirow{3}{*}{REC.}
& Where is $<$object$>$ in the image?  \\
& What are the coordinates for the given $<$object$>$ in the image?    \\
& Given the image, could you please tell me where is $<$object$>$  \\
\cmidrule(lr){1-1}\cmidrule(lr){2-2}
\multirow{3}{*}{Phrase Grounding}
& What are the locations of $<$objects$>$?  \\
& Could you provide me with the exact locations of $<$objects$>$?    \\
& Please indicate the positions of $<$objects$>$ in the image?  \\
\cmidrule(lr){1-1}\cmidrule(lr){2-2}
\multirow{3}{*}{Object Detection (O365)}
& Detect all objects among $<$class$>$ in the image.  \\
& Perform object detection given the image within $<$class$>$.    \\
& Given the image and set $<$class$>$, identify all the objects that belong to the set.  \\
\cmidrule(lr){1-1}\cmidrule(lr){2-2}
\multirow{3}{*}{Grounded Captioning}
& What is this photo about? Use concise language.  \\
& Describe the overall picture in just a few words.  \\
& What do you see happening in this image? Provide the answer in short.  \\
\cmidrule(lr){1-1}\cmidrule(lr){2-2}
\multirow{3}{*}{Object Hallucination}
& Is there a $<$object$>$ in the image? \\
& Are there $<$object$>$ in the image?\\
& Please tell me whether $<$object$>$ exists in the image? \\
\bottomrule
\end{tabular}%
\label{tab:task_temp}
}
\end{table*}

\subsection{Details on Spatial Negative Mining}
\label{app:negative_mining}

In Section~\ref{sec:negative_mining}, we conducted negative sample mining for two aspects: ($i$) \textit{Image-conditioned Category Localization}, and ($ii$) \textit{Semantics-conditioned Category Localization}.  They use the same template to convert the original data, which falls into the task of object hallucination in Table \ref{tab:task_temp}. Specifically, for the negative categories in  ($ii$), we prompt ChatGPT/GPT-4 to generate entities that are most analogous to the original class, attribute, or quantity, \emph{e.g.}, `man' vs. `woman', `blue' vs. `yellow', `two' vs. `three'. The prompt feed into ChatGPT/GPT-4 encompasses all the entities extracted from 5 captions associated with one single image. We show the exact prompt template in Table~\ref{tab:neg_prompt_2}.
% They both ask the model to pinpoint specific object categories, thereby enabling the model's ability to discern and potentially recognize the absence of certain objects.  They differ in how to select the negative category.  For ($i$), Object365 data are employed and we randomly select the object class from the vocabulary that is not shown in the given image.  For ($ii$), Flickr30k data are used and negative categories are sourced by utilizing ChatGPT/GPT4 to pinpoint entities that are most analogous to the original class, attribute, or quantity, \emph{e.g.}, `man' vs. `woman', `blue' vs. `yellow', `two' vs. `three'. 

% One prompt list encompasses all the entities extracted from 5 captions associated with one single image. We present the prompt template utilized for this purpose in Table~\ref{tab:neg_prompt_2}.

% we conducted negative sample mining for two aspects, including \textit{(i) Region-centric Category Verification}, and \textit{(ii) Unconditional Category Localization}. In the realm of (i), the primary objective is to ascertain whether a specific region or object within an image aligns with a designated category. 

% For (ii), the model is tasked with pinpointing specific categories within an image, irrespective of the category's actual presence. The negative sets are derived using ChatGPT/GPT4 by prompting on the Flickr30k Entities dataset. One prompt list encompasses all the entities extracted from 5 captions associated with one single image. We present the prompt template utilized for this purpose in Table~\ref{tab:neg_prompt_2}.

\begin{table*}[t]\centering
\caption{In this example, we provide the prompt to generate the spatial negative sets.}
\begin{minipage}{0.99\columnwidth}\vspace{0mm}    \centering
\begin{tcolorbox} 
    \centering
    \small
     \hspace{-6mm}
    \begin{tabular}{p{0.99\columnwidth}}

\begin{minipage}{0.99\columnwidth}\vspace{0mm}

\VarSty{messages} = [
            \{\var{"role":"system", "content":} f'''You are an AI visual assistant that can analyze a single image. You receive \textbf{several entities} given by a list, each describing the objects in the image you are observing. \\
            
For each entity mentioned, change them with the most misleading entity name (may belong to the same category but are actually different) (\textbf{nonexistent objects}: man $\rightarrow$ woman, \textbf{nonexistent attributes}: brown $\rightarrow$ yellow, \textbf{nonexistent quantities}: two $\rightarrow$ three, etc.). The instructions should contain interrogative and declarative sentences. \\

The output format needs to be a list only which contains the misleading entity names. Please follow the instructions carefully. \\

1. The length of the output list needs to be exactly equal to the input list.\\

2. Do not explain the reasons. \\

3. Do not mention the input entities, at least the output name and input name needs to be different. \\

4. Do not mention something abstract, like \"alien\". \\

5. When dealing with quantities, focus solely on increasing the numbers during revision. \\

6. When dealing with words like "a few", "a group", "several", "some", etc., try changing the objects (A few men $\rightarrow$ A few women). \\

7. Ensure that inclusive words are not substituted with their specific subsets. For example, if the word is "people," avoid replacing it with genders like "man" or "woman." Instead, consider modifying them to different categories, such as "people" $\rightarrow$ "animals.".'''\}]
\end{minipage}
    \end{tabular}
\end{tcolorbox}
    
\vspace{-2mm}
    \label{tab:neg_prompt_2}
\end{minipage}
\end{table*}

\subsection{Examples for Generating Refer-and-Ground datasets}
\label{app:prompt}
We provide some example prompts to generate refer-and-ground from ChatGPT/GPT-4. Prompt and the in-context example of multiple-round visual conversation data are shown in Table~\ref{tab:prompt_conversation} and Table~\ref{tab:example1_cap_conv}. Prompt and the in-context example of one-round reasoning data are shown in Table~\ref{tab:prompt_reason} and Table~\ref{tab:example1_cap_reason}.

\begin{table*}[htbp]\centering
\caption{In this example, we provide the prompt used to generate the conversation response for refer-and-ground instruction tuning, following the practice of LLaVA~\citep{liu2023visual}.}
\begin{minipage}{0.99\columnwidth}\vspace{0mm}    \centering
\begin{tcolorbox} 
    \centering
    \small
     \hspace{-6mm}
    \begin{tabular}{p{0.99\columnwidth}}

\begin{minipage}{0.99\columnwidth}\vspace{0mm}

\VarSty{messages} = [
            \{\var{"role":"system", "content":} f'''You are an AI visual assistant that can analyze a single image. You receive five global captions, each describing the same image you are observing. In addition, specific object locations within the image are given, along with detailed coordinates. These coordinates are in the form of bounding boxes, represented as (x1, y1, x2, y2) with floating numbers ranging from 0 to 1. These values correspond to the top left x, top left y, bottom right x, and bottom right y. Also, the relationships between pairs of objects are provided in the format of object $\rightarrow$ relationship $\rightarrow$ subject, where the object/subject are indexed by object id from previous object lists as well as the object names. Also, several region descriptions are given, each describing a box region of the image, with detailed coordinates. \\
            
Design a conversation between you and a person asking about this photo. Ask diverse questions and give corresponding answers. The answers should be in a tone that a visual AI assistant is seeing the image and answering the question. \\

Here are some additional requirements about generated questions and answers: \\

1. Only include questions that have definite answers:\\
(1) one can see the content in the image that the question asks about and can answer confidently; \\
(2) one can determine confidently from the image that it is not in the image.
Do not ask any questions that cannot be answered confidently. \\

2. Also include complex questions that are relevant to the content in the image, for example, asking about background knowledge of the objects in the image, asking to discuss events happening in the image, asking about object actions in the context of entire images, etc. Again, do not ask about uncertain details. \\

3. Provide detailed answers when answering complex questions. For example, give detailed examples or reasoning steps to make the content more convincing and well-organized.  You can include multiple paragraphs if necessary. \\

4. In all samples, either in question or answer, you must mention bounding box coordinates to refer to the object or regions instead of directly saying the object name or describing the regions in text. In answer, explain the region in the context of the scene. \\

5. Do not mention that the information source is provided in the text/caption/region description.  Always answer as if you are directly looking at the image. \\

6. Make the question as diverse as possible. Include questions asking about the visual content of the image, including the object types, counting the objects, object actions, object locations, relative positions between objects, object selection, object functions, etc. Make the question challenging by less including the visual content details in the question.'''\}\\
        ]
        
    \For{ \VarSty{sample} in   \VarSty{fewshot\_samples}}{
         \var{\VarSty{messages}.append(\{"role":"user", "content":\VarSty{sample[`context']}\})} \; \\
         \var{\VarSty{messages}.append(\{"role":"assistant", "content":\VarSty{sample[`response']}\} ) } \;
         }  
    \var{\VarSty{messages}.append(\{"role":"user", "content":`\textbackslash  n'.join(\VarSty{query})\})}
\end{minipage}
    \end{tabular}
\end{tcolorbox}
    
\vspace{-2mm}
    \label{tab:prompt_conversation}
\end{minipage}
\end{table*}

\begin{table*}[htbp]\centering
\caption{One example used in in-context learning to construct GPT-Assisted Refer-and-Ground Instructon-Tuning. We encourage readers to refer to the codebase for the prompts for other types.}
\begin{minipage}{0.99\columnwidth}\vspace{0mm}    \centering
\begin{tcolorbox} 
    \centering
   
     \hspace{-4mm}
      \scriptsize
    \begin{tabular}{p{0.99\columnwidth} c} 
    \VarSty{ {\bf Objects} } 
    & \hspace{-6.1cm} \multirow{4}{*}{ \includegraphics[height=4.7cm]{figs/prompt_temp1.jpg} } \\
    Object 0: chair at [0.596, 0.637, 0.698, 0.997]. \\
    \hspace{0.5cm}... \\
     \VarSty{ {\bf Relationships} } \\
    Object 11 : frame $\rightarrow$ with $\rightarrow$ Object 9 : photo \\
    \hspace{0.5cm}... \\
    \VarSty{ {\bf Region Descriptions} } \\
    Region Description at [0.560, 0.466, 0.600, 0.529] : \\
    \hspace{1cm} a white picture frame with a black and white photo on it. \\
    \hspace{0.5cm}... \\
    \VarSty{ {\bf Global Caption} } \\
    White chairs sit around a polished wood dining table while \\
    \hspace{1cm} a sectional soft sits in the background. \\
    \hspace{0.5cm}... \\
\\
    % \midrule
   \VarSty{ {\bf Conversation} } \\
Question:  \\
What is the object [0.212, 0.539, 0.718, 0.820] made of? \\
=== \\
Answer: \\
The dining table over there is crafted from a tree trunk. \\
=== \\
Question: \\ 
Where would be a good place to have my dinner? Point out the locations in image.\\
=== \\
Answer: \\
I suggest having dinner at the table  [0.214, 0.541, 0.720, 0.997], possibly sitting on the white chair [0.264, 0.717, 0.524, 0.995]. Also, enjoy the wine [0.242, 0.644, 0.288, 0.682]. \\
=== \\
Question: \\
In terms of comfort for resting, which object is better - the one  [0.830, 0.431, 0.978, 0.596] or the one [0.264, 0.717, 0.524, 0.995]? \\
=== \\
Answer: \\
Among the two, the couch [0.830, 0.431, 0.978, 0.596] would be more comfortable for resting. The object  [0.264, 0.717, 0.524, 0.995] is a dining chair, which is typically less comfortable for extended relaxation. \\
=== \\
Question: \\
Is the item  [0.182, 0.238, 0.342, 0.461] real or is it a mirror reflection? \\
=== \\
Answer: \\
The television and the fireplace [0.182, 0.238, 0.342, 0.461] are not real but are actually reflections in the mirror.  \\
=== \\
Question: \\
Can you describe the decorating style of the living room shown in this image? \\
=== \\
Answer: \\
The living room is decorated in a modern chic style. The image shows white chairs and lamps, a gray couch, and a wooden table and floor, all contributing to a modern aesthetic. The large mirror and built-in wine rack are also characteristic of modern design. Thus, the overall decor creates an atmosphere of elegance and sophistication, indicating a modern chic style.

    \end{tabular}
\end{tcolorbox}
\vspace{-2mm}
    \label{tab:example1_cap_conv}
\end{minipage}
\end{table*}

\begin{table*}[htbp]\centering
\caption{In this example, we provide the prompt used to generate the reasoning response for refer-and-ground instruction tuning, following the practice of LLaVA~\citep{liu2023visual}.}
\begin{minipage}{0.99\columnwidth}\vspace{0mm}    \centering
\begin{tcolorbox} 
    \centering
    \small
     \hspace{-6mm}
    \begin{tabular}{p{0.99\columnwidth}}

\begin{minipage}{0.99\columnwidth}\vspace{0mm}

\VarSty{messages} = [
            \{\var{"role":"system", "content":} f'''You are an AI visual assistant that can analyze a single image. You receive five global captions, each describing the same image you are observing. In addition, specific object locations within the image are given, along with detailed coordinates. These coordinates are in the form of bounding boxes, represented as (x1, y1, x2, y2) with floating numbers ranging from 0 to 1. These values correspond to the top left x, top left y, bottom right x, and bottom right y. Also, the relationships between pairs of objects are provided, in the format of object $\rightarrow$ relationship $\rightarrow$ subject, where the object/subject are indexed by object id from previous object lists as well as the object names. Also, several region descriptions are given, each describing a box region of the image, with detailed coordinates. \\
            
The task is to use the provided image information (objects, attribute, relationship, region description, captions), create a plausible and challenging question about the image, and provide the answer in detail. \\

Create complex questions that mention specific regions of the image, but the question should require some knowledge-aware or high-level commonsense reasoning beyond describing the scene. \\

To answer such questions, one should first understand the visual content, then based on the background knowledge or reasoning, either explain why the things are happening that way or provide guides and help to the user's request.  Make the question challenging by not including the visual content details in the question so that the user needs to reason about that first. \\

Here are some additional requirements about generated questions and answers: \\

1. In question or answer, you must mention bounding box coordinates to refer to the object or regions, instead of directly say the object name or describing the regions in text.  In answers, explain the region in the context of scene. Include details like object counts, position of the objects, relative position between the objects.  \\

2. Don't ask the question you are not confident to answer.  Only include question that have definite answer. \\

3. Do not mention that the information source is provided in text/catpion/region description.  Always answer as if you are directly looking at the image. \\

4. Make the question as diverse as possible and as complex-reasoning required as possible.'''\}\\
        ]
        
    \For{ \VarSty{sample} in   \VarSty{fewshot\_samples}}{
         \var{\VarSty{messages}.append(\{"role":"user", "content":\VarSty{sample[`context']}\})} \; \\
         \var{\VarSty{messages}.append(\{"role":"assistant", "content":\VarSty{sample[`response']}\} ) } \;
         }  
    \var{\VarSty{messages}.append(\{"role":"user", "content":`\textbackslash  n'.join(\VarSty{query})\})}
\end{minipage}
    \end{tabular}
\end{tcolorbox}
    
\vspace{-2mm}
    \label{tab:prompt_reason}
\end{minipage}
\end{table*}

\begin{table*}[htbp]\centering
\caption{One example used in in-context learning to construct GPT-Assisted Refer-and-Ground Instructon-Tuning.  We encourage readers to refer to the codebase for the prompts for other types.}
\begin{minipage}{0.99\columnwidth}\vspace{0mm}    \centering
\begin{tcolorbox} 
    \centering
   
     \hspace{-4mm}
      \scriptsize
    \begin{tabular}{p{0.99\columnwidth} c}
    \VarSty{ {\bf Objects} } 
    & \hspace{-5.3cm} \multirow{4}{*}{ \includegraphics[height=4.3cm]{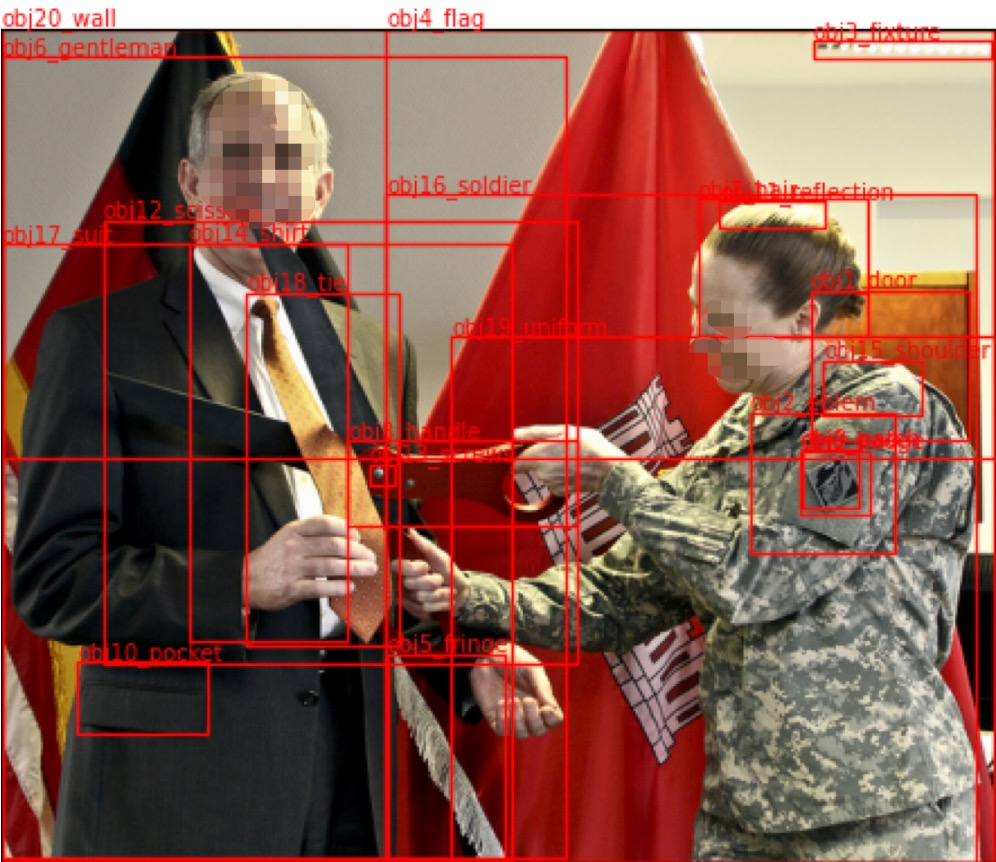} } \\
    Object 0: badge at [0.802, 0.505, 0.872, 0.581]. \\
    \hspace{0.5cm}... \\
     \VarSty{ {\bf Relationships} } \\
    Object 5 : fringe $\rightarrow$ on $\rightarrow$ Object 4 : flag \\
    \hspace{0.5cm}... \\
    \VarSty{ {\bf Region Descriptions} } \\
    Region Description at [0.038, 0.029, 0.508, 0.995] :\\
    A Grinning gentlemen about to get his tie out. \\
    \hspace{0.5cm}... \\
    \VarSty{ {\bf Global Caption} } \\
    A female in military uniform cutting a businessman’s neck tie.  \\
    \hspace{0.5cm}... \\
\\
    % \midrule
   \VarSty{ {\bf Conversation} } \\
Question:  \\
What is unusual about the region [0.102, 0.230, 0.578, 0.761]?
 \\
=== \\
Answer: \\
In the specified region, there is an intriguing sight: a pair of oversized scissors positioned near a man's tie, accompanied by a woman in military attire. The two of them share a hearty laugh, creating an unexpected contrast. The atmosphere appears celebratory, resembling a ribbon-cutting ceremony with flags in the background. However, rather than cutting a ribbon, the scissors are directed toward the man's tie, adding an unexpected twist to the situation. \\
=== \\

    \end{tabular}
\end{tcolorbox}
\vspace{-2mm}
    \label{tab:example1_cap_reason}
\end{minipage}
\end{table*}

\newpage
\section{Examples and Prompts For Ferret-Bench}
\label{app:ferretbench}

We leverage GPT-4 to generate three kinds of region-based questions evaluating referring and grounding capability: (i) Referring Description, (ii) Referring Reasoning, and (iii) Grounding in Conversation. Here, we only provide the prompt in Table~\ref{app:prompt_refer_desc} used to generate the referring description response. One example of GPT-4 answers is shown in Table~\ref{app:example_refer_desc}. We recommend readers check out more examples in Appendix~\ref{app:viz}.

\begin{table*}[htbp]\centering
\caption{In this example, we provide the prompt used to generate the referring description response.}
\begin{minipage}{0.99\columnwidth}\vspace{0mm}    \centering
\begin{tcolorbox} 
    \centering
    \small
     \hspace{-6mm}
    \begin{tabular}{p{0.99\columnwidth}}

\begin{minipage}{0.99\columnwidth}\vspace{0mm}

\VarSty{messages} = [
            \{\var{"role":"system", "content":} f'''You are an AI visual assistant that can analyze a single image. You receive five global captions, each describing the same image you are observing. In addition, specific object locations within the image are given, along with detailed coordinates. These coordinates are in the form of bounding boxes, represented as (x1, y1, x2, y2) with floating numbers ranging from 0 to 1. These values correspond to the top left x, top left y, bottom right x, and bottom right y. Also, the relationships between pairs of objects are provided, in the format of object $\rightarrow$ relationship $\rightarrow$ subject, where the object/subject are indexed by object id from previous object lists as well as the object names. Also, several region description are given, each describing a box region of image, with detailed coordinates.  \\
            
The task is to use the provided image information (objects, attribute, relationship, region description, captions), create a plausible and challenging question about the image, and provide the answer in detail. \\

Create questions that refer to coordinates of some objects or regions without describing it, and ask about its interaction with surrounding/nearby objects. \\

To answer such questions, one should require first understanding the visual content, then based on the spatial information provided. \\

Here are some additional requirements about generated questions and answers: \\

1. In question, you must mention bounding box coordinates to refer to the object or regions, instead of directly say the object name or describing the regions in text.  In answers, explain the region in the context of scene. Include details like object counts, position of the objects, relative position between the objects.    \\

2. Don't ask the question you are not confident to answer.  Only include question that have definite answer.  \\

3. Do not mention that the information source is provided in text/catpion/region description.  Always answer as if you are directly looking at the image. \\

4. Don't mention additional coordinates in the answer. \\

5. Question should be explicitly ask about context/surrounding/nearby information/interaction.'''\}\\
        ]
        
    \For{ \VarSty{sample} in   \VarSty{fewshot\_samples}}{
         \var{\VarSty{messages}.append(\{"role":"user", "content":\VarSty{sample[`context']}\})} \; \\
         \var{\VarSty{messages}.append(\{"role":"assistant", "content":\VarSty{sample[`response']}\} ) } \;
         }  
    \var{\VarSty{messages}.append(\{"role":"user", "content":`\textbackslash  n'.join(\VarSty{query})\})}
\end{minipage}
    \end{tabular}
\end{tcolorbox}
    
\vspace{-2mm}
    \label{app:prompt_refer_desc}
\end{minipage}
\end{table*}

\begin{table*}[htbp]\centering
\caption{One example of GPT4's answers about referring description. }
\begin{minipage}{0.99\columnwidth}\vspace{0mm}    \centering
\begin{tcolorbox} 
    \centering
   
     \hspace{-4mm}
      \scriptsize
    \begin{tabular}{p{0.99\columnwidth}}
   \VarSty{ {\bf Conversation} } \\
Question:  \\
Describe the region [0.102, 0.230, 0.578, 0.761] in the context of surrounding objects.
 \\
Answer: \\

It's a man in the region, who is standing on top of a skateboard, performing a rail grind over a metal hand rail. There is a bench and a tree near him. Also, he is in front of some buildings. \\
\centering
\vspace{0.3cm}
\includegraphics[height=5.2cm]{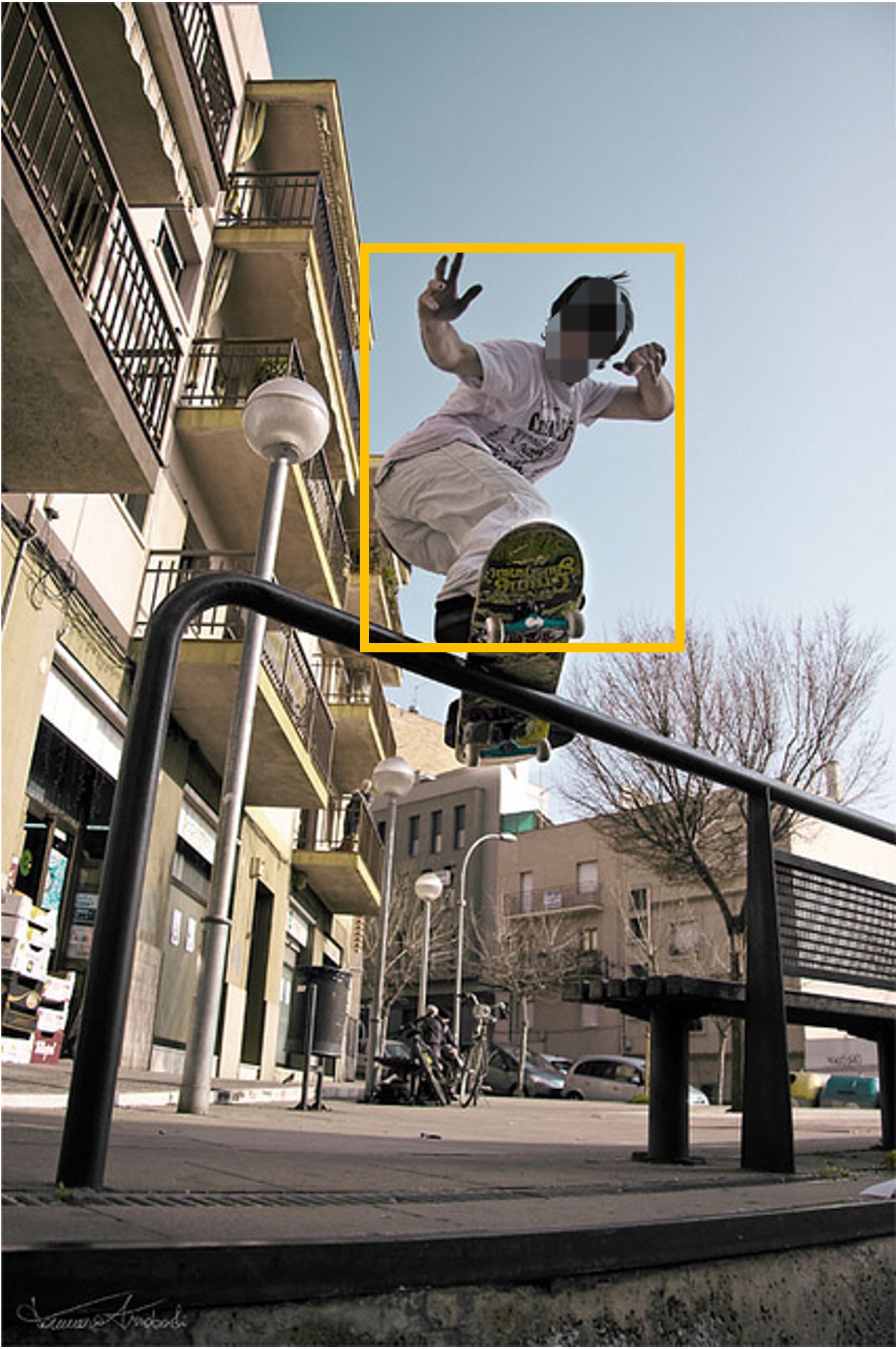}

    \end{tabular}
\end{tcolorbox}
\vspace{-2mm}
    \label{app:example_refer_desc}
\end{minipage}
\end{table*}

\newpage
\section{More Visualization}
\label{app:viz}
\begin{figure}[htbp]
\centering
\includegraphics[width=1.1\linewidth]{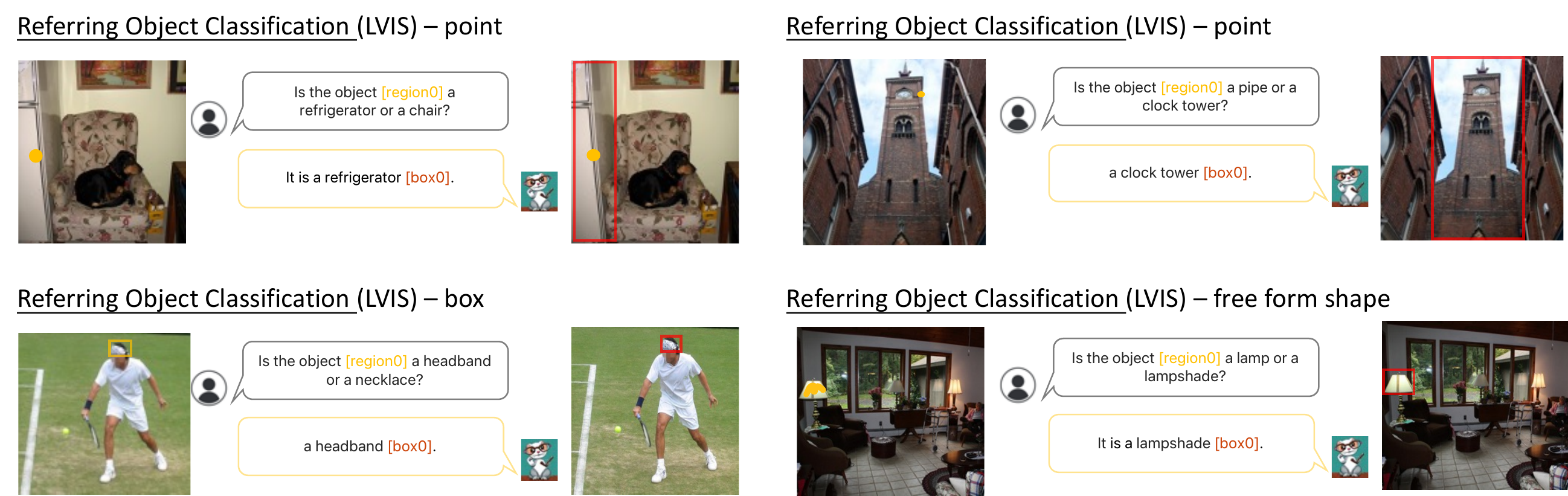} 
\caption{\textbf{Referring Object Classification on LVIS}. The task aims to classify specific region(s) in an image given by point/box/segmentation inputs.}
% \haotian{add more descriptions about point/box}}
\label{fig:lvis}
\end{figure}

\begin{figure}[htbp]
\centering
\includegraphics[width=1.1\linewidth]{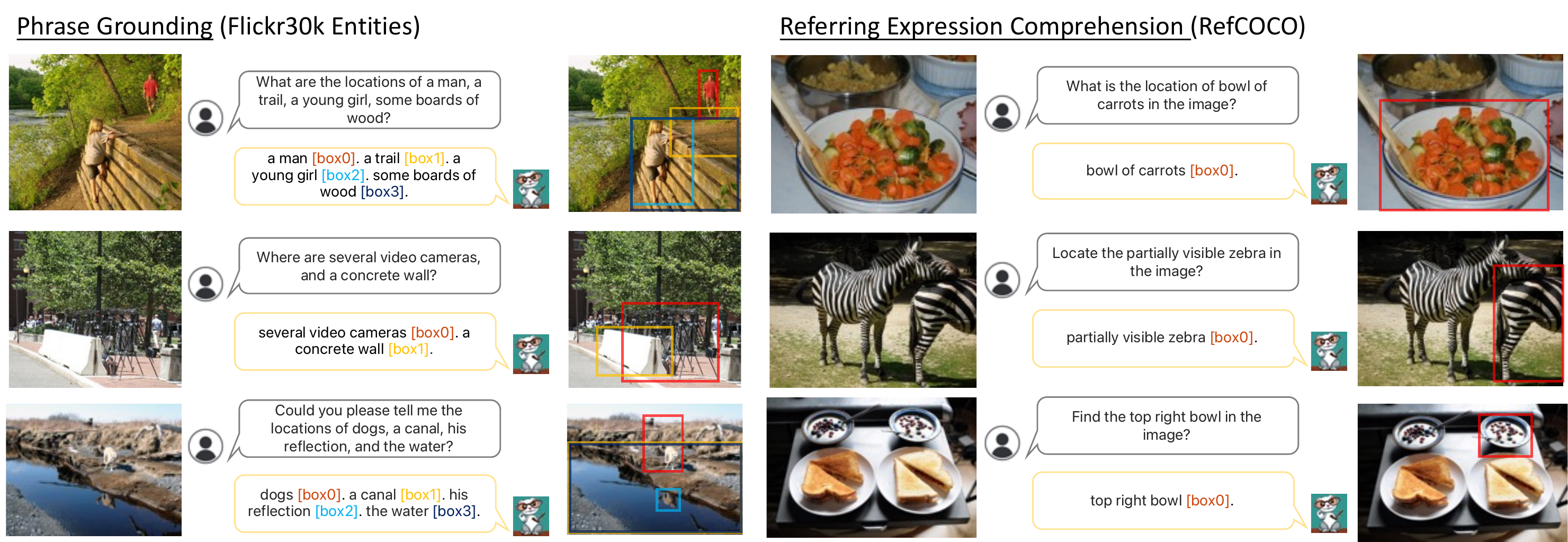}

\vspace{0.3cm}

\includegraphics[width=1.1\linewidth]{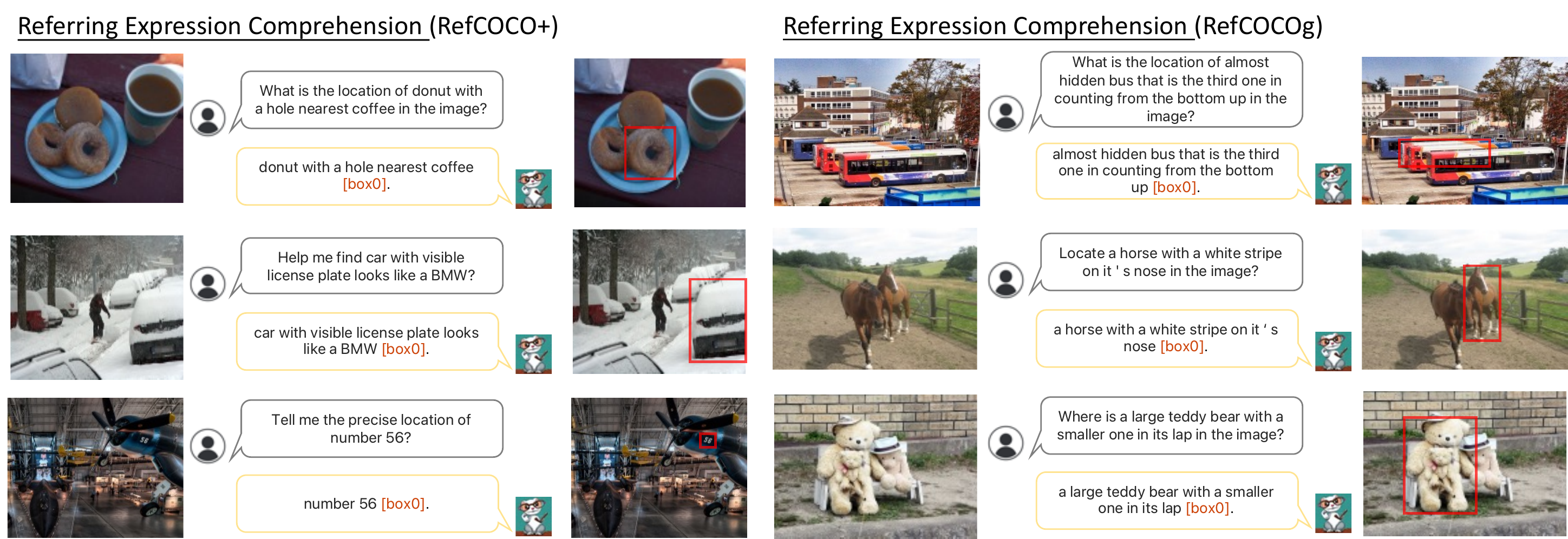}
\caption{\textbf{Phrase Grounding} on Flickr30k Entities and \textbf{Referring Expression Comprehension} on RefCOCO/RefCOCO+/RefCOCOg. The tasks aim to localize specific object(s) in an image described by a referring expression/given entity.}
\label{fig:visual_ground}
\end{figure}

\begin{figure}[htbp]
\centering
\includegraphics[width=1.1\linewidth]{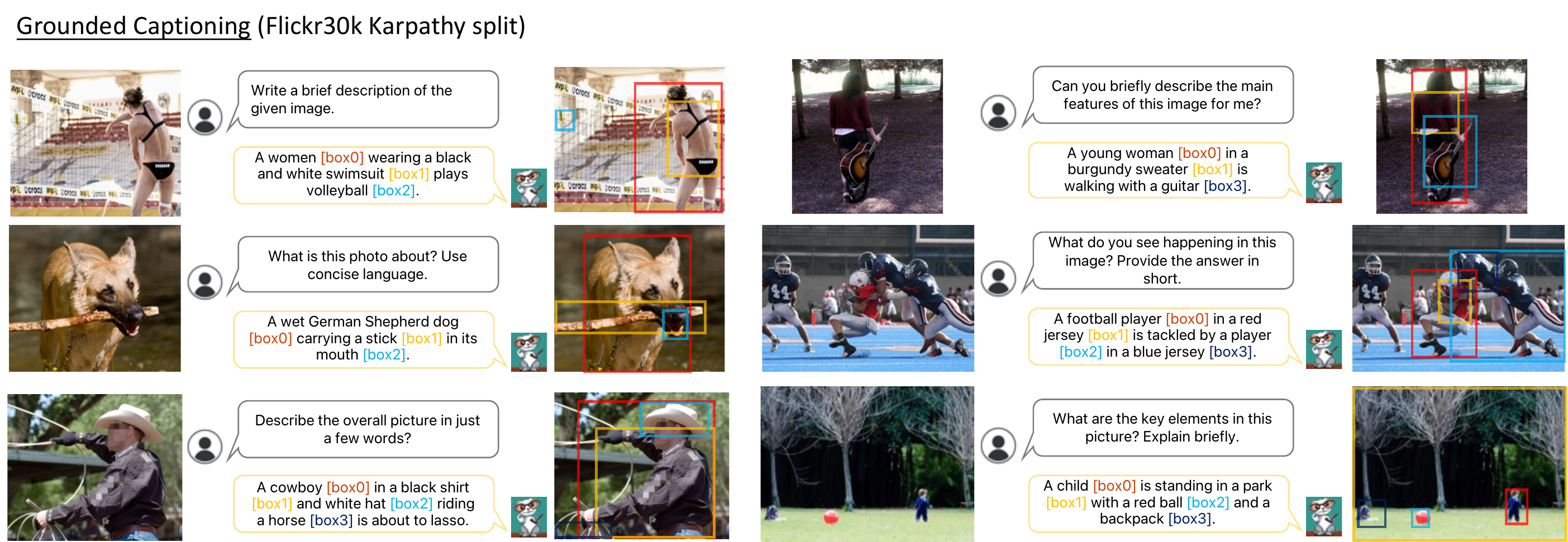} 
\caption{\textbf{Grounded Captioning on Flickr30k}. The task aims to generate a caption about the image and ground all generated noun phrases to image regions.}
\label{fig:grounded_caption}
\end{figure}

\begin{figure}[htbp]
\centering
\includegraphics[width=1.1\linewidth]{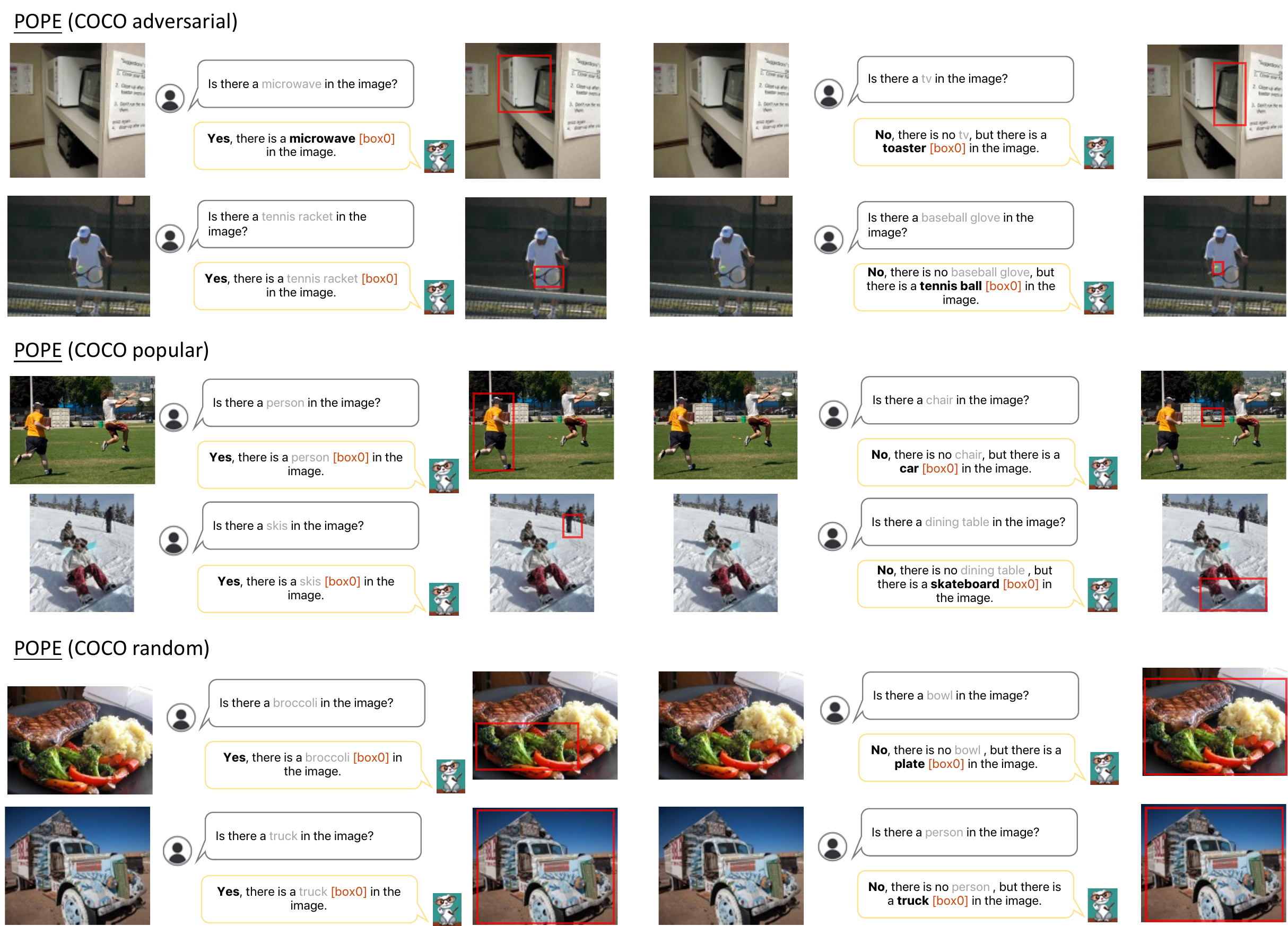} 
\caption{\textbf{Object Hallucination Evaluation (POPE) on COCO}. The task aims to input a query inquiring about the existence of an object, and the model is expected to generate a response in the form of either ``yes/no''.}
\label{fig:pope}
\end{figure}

We provide more quantitative results of the predictions under various tasks from \ferret to indicate the model's strength and capability. 
\begin{itemize}[leftmargin=*]
 \item Please refer to Figure~\ref{fig:lvis} for Referring Object Classification on LVIS with different referring formats (point/box/).
 \item Please refer to Figure~\ref{fig:visual_ground} for Visual Grounding on Flickr30k Entities and Referring Expression Comprehension on RefCOCO/RefCOCO+/RefCOCOg. 
 \item Please refer to Figure~\ref{fig:grounded_caption} for Grounded Captioning on Flickr30k Karpathy split. 
 \item Please refer to Figure~\ref{fig:pope} for Evaluating Object Hallucination (POPE) on COCO val split. 
 \item Please refer to Table~\ref{tab:refer_descip} for Referring Description in Ferret-Bench.
 \item Please refer to Table~\ref{tab:refer_reason} for Referring Resoning in Ferret-Bench.
 \item Please refer to Table~\ref{tab:ground_conv} for Grounding in Conversation in Ferret-Bench.
\end{itemize}

\begin{table*}[t!]\centering
\caption{\textbf{Referring Description in Ferret-Bench.} Qualitative examples to illustrate the difference between various models (LLaVA vs. Kosmos-2 vs. Shikra vs. Ferret (Ours)). For clarity, we have omitted the bounding box outputs in the textual representations.}
\begin{minipage}{1.0\columnwidth}\vspace{0mm}    \centering
\begin{tcolorbox} 
    \centering
    %  \hspace{-10mm}
      \footnotesize
    \begin{tabular}{p{0.97\columnwidth} c}
   % \VarSty{ {\bf Input} } & \\
\ArgSty{\bf Question:}  
& \hspace{-5.3cm} \multirow{4}{*}{ \includegraphics[height=3.8cm]{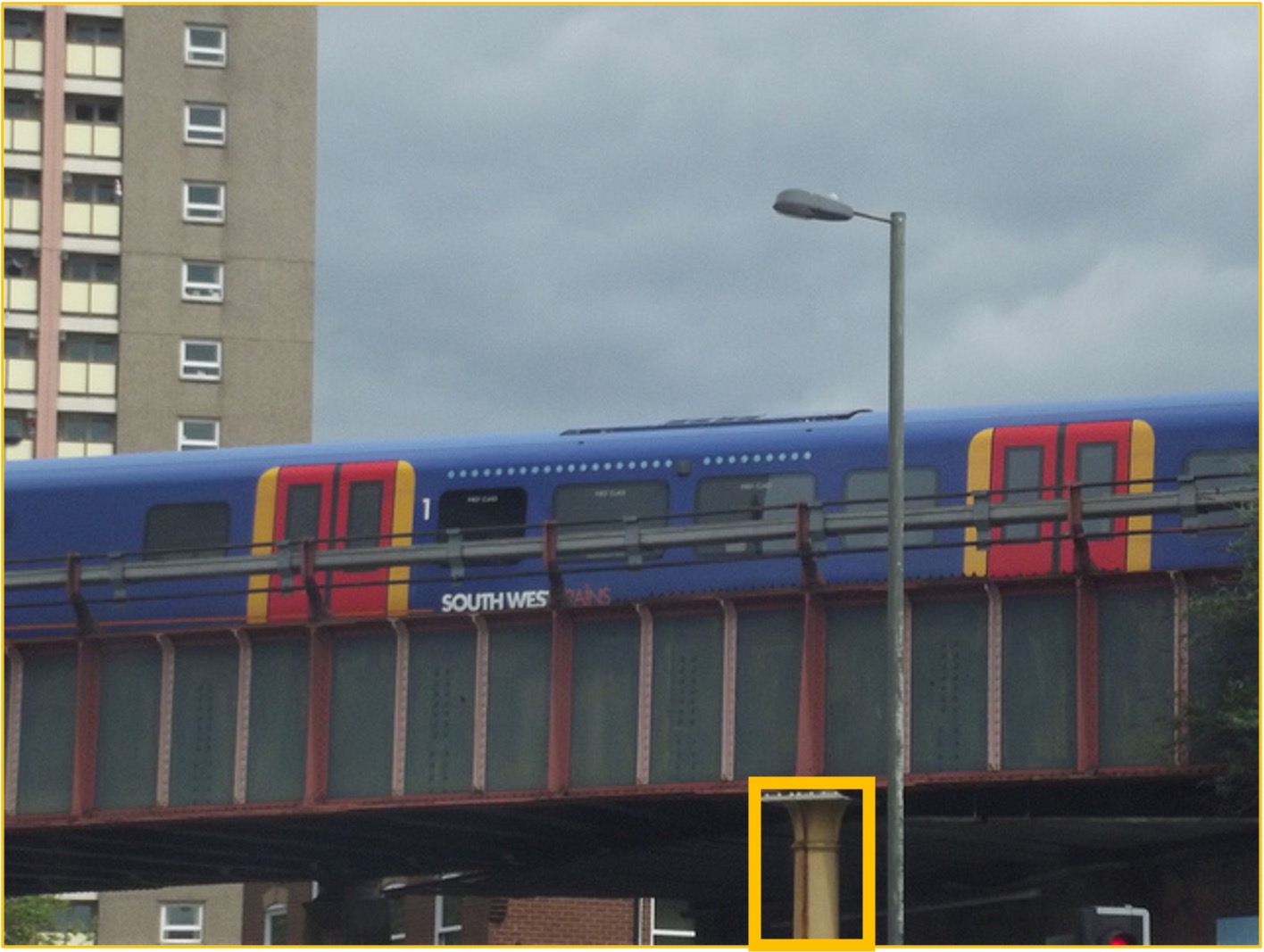} } \\
What is the interaction between the object \textcolor{orange}{[region0]} &\\
and its surrounding objects?\\
 \\
 % \\
\CommentSty{\bf Ground Truth:} \\ 
In the region, there is a metal support column. &\\ 
This column is providing support for a bridge above it, &\\
which a train is passing over.  The column also features &\\ 
a red line on it. This region is part of a larger scene that &\\
includes a train track on an elevated bridge. \\
    & \\
   \VarSty{ {\bf LLaVA:} } & \\
The object is a bridge which appears to be a part of a train station, as there are traffic lights present in the scene. The traffic lights are likely used to control the train's movement and ensure safe passage over the bridge.  \\
   \VarSty{ {\bf Kosmos-2}: }& \\
The interaction between the object and its surroundings is that of a train passing over a bridge.  \\
   \VarSty{ {\bf Shikra}: }& \\
The metal pole is standing tall and straight, supporting the train that is traveling on the elevated track. \\
   \VarSty{ {\bf \ferret (Ours)}: }& \\
    The object is a metal post that is part of a bridge. This post is essential because it supports the bridge structure, ensuring its stability for the train to safely traverse the bridge. Other elements such as the train, and the overhead guard rail, depend on this post for support and stability. Without this post, the bridge wouldn't be able to withstand the train's weight and motion, potentially leading to a structural collapse. Therefore, this post plays a critical role in maintaining the bridge's integrity and stability. \\
    \hrulefill 
    \\
    \\
    \ArgSty{\bf Question:}  
& \hspace{-5.3cm} \multirow{4}{*}{ \includegraphics[height=5cm]{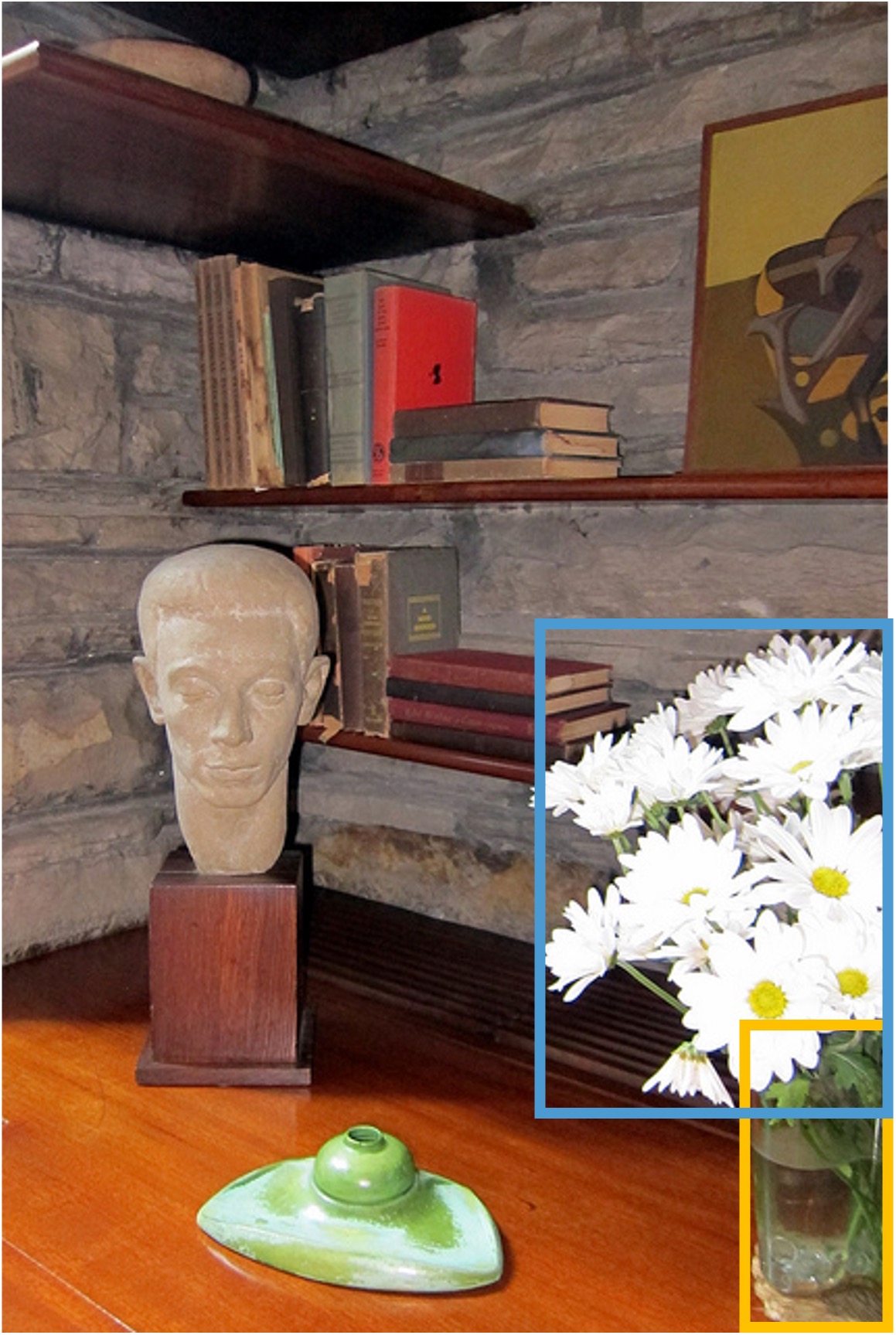} } \\
What is the relationship between the object \textcolor{orange}{[region0]} & \\
and object \textcolor{blue}{[region1]} and its surrounding objects? \\
\\
 % \\
\CommentSty{\bf Ground Truth:} \\ 
The object is a vase, and the object is a flower. &\\
The flower is in the vase, suggesting it is a &\\
decorative element within the room.  \\
\\
\VarSty{ {\bf LLaVA:} } & \\
A wooden table with a green bowl placed on it.  \\
   \VarSty{ {\bf Kosmos-2}: }& \\
The object is a vase with a green base. \\
   \VarSty{ {\bf Shikra}: }& \\
The answer is flowers. \\
   \VarSty{ {\bf \ferret (Ours)}: }& \\
    The object and object are a vase and a flower, respectively. The vase is holding the flower, suggesting it's displaying or supporting it. \\
    \end{tabular}
\end{tcolorbox}
\vspace{-2mm}
    \label{tab:refer_descip}
\end{minipage}
\end{table*}

\begin{table*}[t!]\centering
\caption{\textbf{Referring Reasoning in Ferret-Bench.} Qualitative examples to illustrate the difference between various models (LLaVA vs. Kosmos-2 vs. Shikra vs. Ferret (Ours)). For clarity, we have omitted the bounding box outputs in the textual representations.}
\begin{minipage}{1.0\columnwidth}\vspace{0mm}    \centering
\begin{tcolorbox} 
    \centering
    %  \hspace{-10mm}
      \footnotesize
    \begin{tabular}{p{0.97\columnwidth} c}
   % \VarSty{ {\bf Input} } & \\
\ArgSty{\bf Question:}  
& \hspace{-5.3cm} \multirow{4}{*}{ \includegraphics[height=3.5cm]{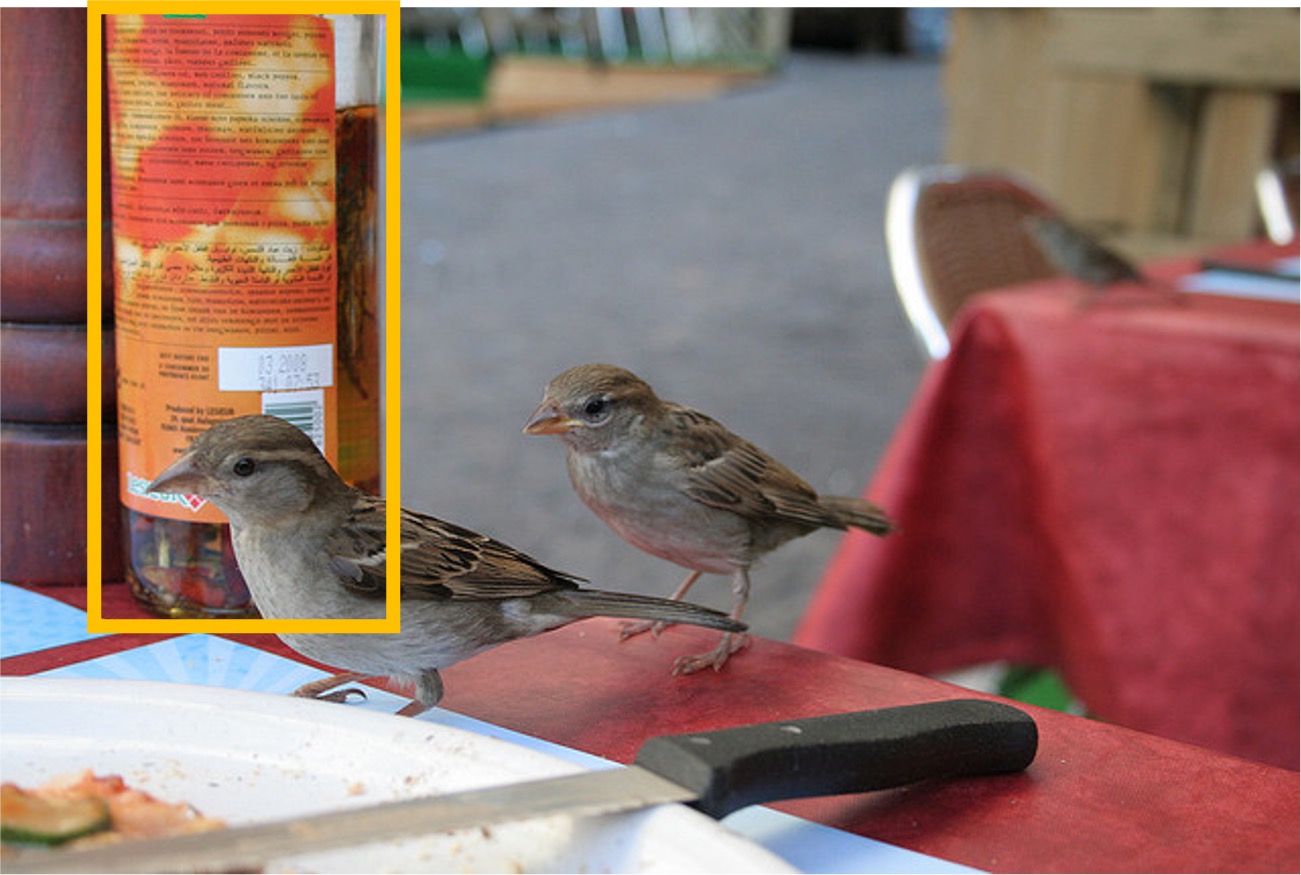} } \\
What is the purpose of the object \textcolor{orange}{[region0]}?\\
 \\
 % \\
\CommentSty{\bf Ground Truth:} \\ 
The object is a bottle. Bottles are typically used to &\\ 
hold and store different types of liquids. In this context, &\\ 
it might be used to store a beverage for the meal. \\
    & \\
   \VarSty{ {\bf LLaVA:} } & \\
A knife and a fork, indicating that it is ready for a meal.  \\
   \VarSty{ {\bf Kosmos-2}: }& \\
The purpose of the object is to attract the birds to the table. \\
   \VarSty{ {\bf Shikra}: }& \\
The answer is keep birds away. \\
   \VarSty{ {\bf \ferret (Ours)}: }& \\
    The object is a bottle, generally used to store and dispense liquids like water, juice, or other beverages. \\
    \hrulefill 
    \\
    \\
    \ArgSty{\bf Question:}  
& \hspace{-5.3cm} \multirow{4}{*}{ \includegraphics[height=3.8cm]{figs/reason_2_mosaic.png} } \\
What is the purpose of the object \textcolor{orange}{[region0]} on the bike?  \\
\\
 % \\
\CommentSty{\bf Ground Truth:} \\ 
The object is a shock absorber on the bike. &\\ 
Its purpose is to absorb or dampen shock impulses.&\\  
It does this by converting the kinetic energy of &\\ 
the shock into another form of energy which is then &\\ 
dissipated. In the context of the motorcycle, &\\ 
it is particularly useful in providing comfort and &\\ 
stability for the rider, especially when travelling &\\ 
over uneven or rough terrains. \\
\\
\VarSty{ {\bf LLaVA:} } & \\
A black and silver Kawasaki motorcycle which is likely used for transportation, commuting, leisure rides, or long-distance travel. \\
   \VarSty{ {\bf Kosmos-2}: }& \\
The purpose of the object on the motorcycle is to store and protect the motorcycle's fuel tank. \\
   \VarSty{ {\bf Shikra}: }& \\
The answer is leg. \\
   \VarSty{ {\bf \ferret (Ours)}: }& \\
    The object is a shock absorber. This component is essential for the bike's stability, providing a smoother ride by soaking up the impact from bumps and uneven surfaces. It significantly improves the overall performance and comfort of the bike, enabling the rider to maintain control and stability even on rough terrains. \\
    \end{tabular}
\end{tcolorbox}
\vspace{-2mm}
    \label{tab:refer_reason}
\end{minipage}
\end{table*}

\begin{table*}[t!]\centering
\caption{\textbf{Grounding in Conversation in Ferret-Bench.} Qualitative examples to illustrate the difference between various models (Kosmos-2 vs. Shikra vs. Ferret (Ours)). For clarity, we only display the bounding box outputs in \ferret's responses.}
\begin{minipage}{1.0\columnwidth}\vspace{0mm}    \centering
\begin{tcolorbox} 
    \centering
    %  \hspace{-10mm}
      \footnotesize
    \begin{tabular}{p{0.97\columnwidth} c}
   % \VarSty{ {\bf Input} } & \\
\ArgSty{\bf Question:}  
& \hspace{-5.3cm} \multirow{4}{*}{ \includegraphics[height=5.6cm]{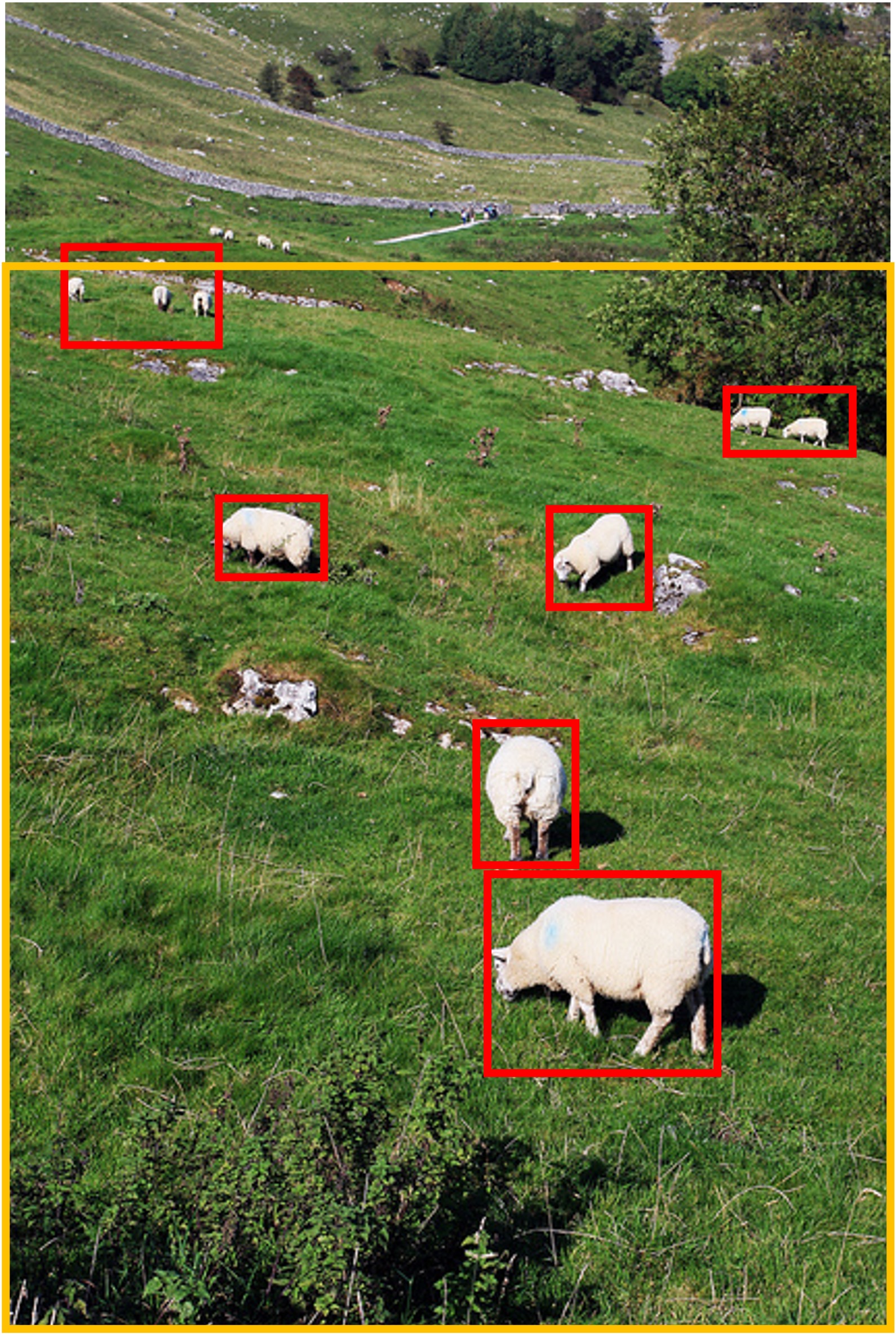} } \\
What is happening in the field and &\\ 
provide the coordinates of mentioned objects?\\
 \\
 % \\
\CommentSty{\bf Ground Truth:} \\ 
In the image, there is a group of sheep [box0] [box1] &\\ 
grazing in a field [box2]. The field is filled with &\\
plant life [box3]. There is also a bush [box4] and &\\
some trees [box5] present in the field. \\
    & \\
   \VarSty{ {\bf Kosmos-2}: }& \\
In the field, a herd of sheep [box0] is grazing on the &\\
lush green grass [box1]. \\
   \VarSty{ {\bf Shikra}: }& \\
A group of sheep [box0] [box1] [box2] [box3] [box4] [box5] & \\
grazing on lush green grass. \\
   \VarSty{ {\bf \ferret (Ours)}: }& \\
    In the field, there are sheep \textcolor{red}{[box0] [box1] [box2] [box3] [box4] [box5]} grazing on the grass \textcolor{orange}{[box6]}. \\
    \hrulefill 
    \\
    \\
    \ArgSty{\bf Question:}  
& \hspace{-5.3cm} \multirow{4}{*}{ \includegraphics[height=5cm]{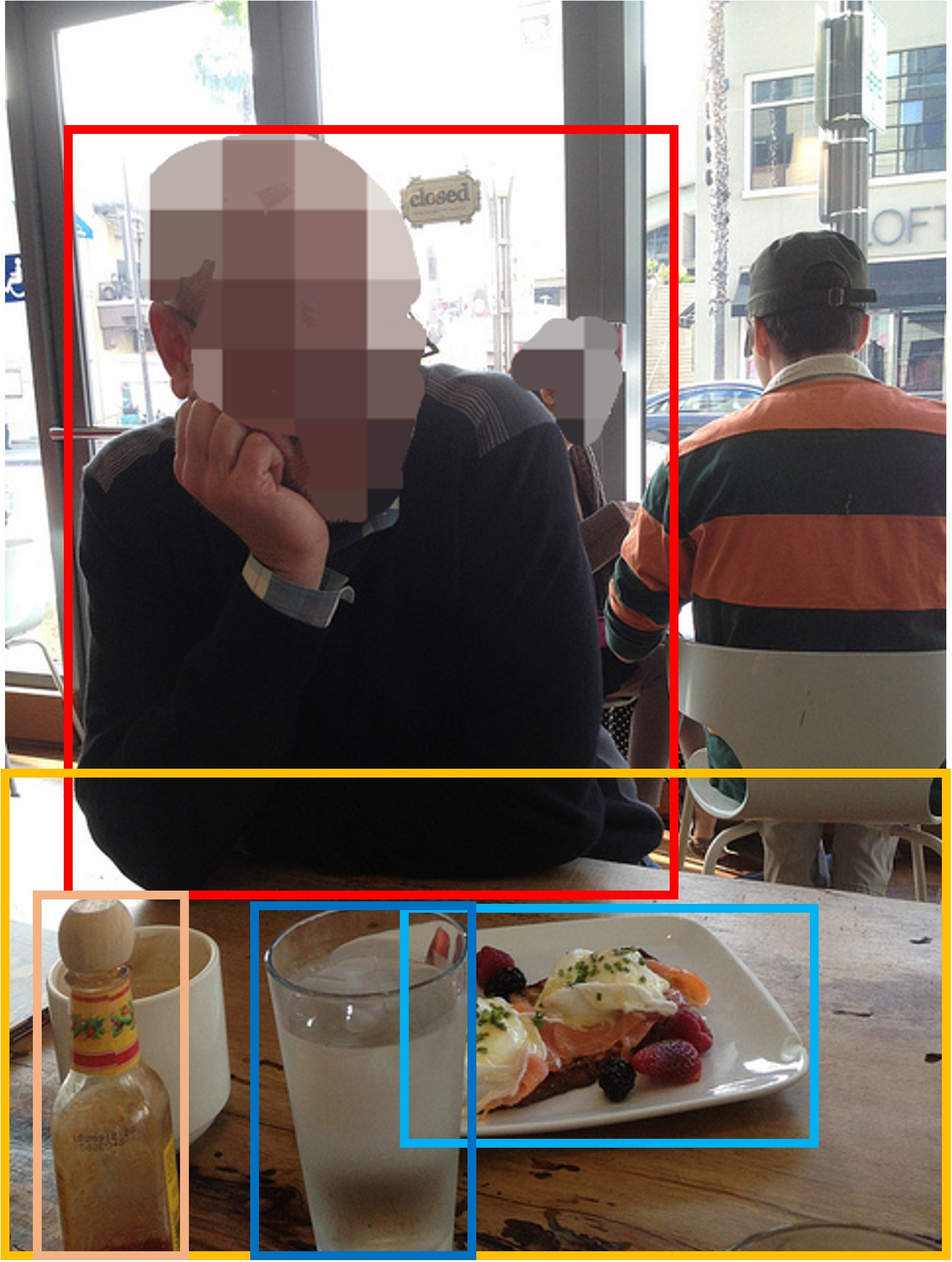} } \\
What is the man doing and what objects are in front of him? & \\
Please provide the coordinates of these objects.  \\
\\
 % \\
\CommentSty{\bf Ground Truth:} \\ 
The man [box0] is sitting at a table [box1] and looking &\\ 
at a plate of food [box2] . There is a glass [box3] &\\ 
and a bottle [box4] on the table in front of him. \\
\\
   \VarSty{ {\bf Kosmos-2}: }& \\
The man is sitting at a dining table [box0] smiling and & \\
looking at the plate of food in front him [box1]. \\
   \VarSty{ {\bf Shikra}: }& \\
The man [box0] is sitting at a table [box1] with a plate & \\
of food [box2] in front of him. He is looking up and & \\
smiling, possibly staring at a person [box0] in the restaurant \\
   \VarSty{ {\bf \ferret (Ours)}: }& \\
    The man \textcolor{red}{[box0]} is having a meal at a table \textcolor{orange}{[box1]}. In front of him, there is a plate \textcolor{blue}{[box2]} with a variety of food items such as fruit and eggs. There is also a glass \textcolor{purple}{[box3]} filled with water and a bottle \textcolor{brown}{[box4]}. \\
    \end{tabular}
\end{tcolorbox}
\vspace{-2mm}
    \label{tab:ground_conv}
\end{minipage}
\end{table*}

% \section{Comparison with GPT-4V on Referring and Grounding Capability}